\documentclass{article} 
\usepackage{collas2024_conference,times}
\usepackage{easyReview}

\usepackage{amsmath,amsfonts,bm,amsthm,amssymb,mathtools}

\def\eqref#1{equation~\ref{#1}}

\def\1{\bm{1}}

\def\vu{{\bm{u}}}

\def\vx{{\bm{x}}}

\def\vz{{\bm{z}}}

\DeclareMathAlphabet{\mathsfit}{\encodingdefault}{\sfdefault}{m}{sl}
\SetMathAlphabet{\mathsfit}{bold}{\encodingdefault}{\sfdefault}{bx}{n}

\newcommand{\E}{\mathbb{E}}

\DeclareMathOperator*{\argmax}{arg\,max}
\DeclareMathOperator*{\argmin}{arg\,min}

\usepackage{hyperref}
\hypersetup{
    colorlinks=true,
    linkcolor=red,
    filecolor=magenta,
    urlcolor=blue,
    citecolor=purple,
    pdftitle={Overleaf Example},
    pdfpagemode=FullScreen,
    }

\usepackage{microtype}
\usepackage{graphicx}
\usepackage{subfigure}
\usepackage{caption} 
\usepackage{wrapfig}
\usepackage{booktabs} 

\usepackage{tabularx}
\usepackage{multirow}

\usepackage{siunitx}
\usepackage{etoolbox}

\usepackage{makecell}

\theoremstyle{plain}
\newtheorem{theorem}{Theorem}[section]

\newtheorem{lemma}[theorem]{Lemma}

\theoremstyle{definition}

\theoremstyle{remark}

\DeclareMathOperator{\disc}{disc}
\DeclareMathOperator{\magn}{mag}

\DeclarePairedDelimiter\abs{\lvert}{\rvert}

\newcommand{\allsource}{\textsc{All Sources}}
\newcommand{\closest}{\textsc{Closest}}

\newcommand{\gft}{\textsc{Gft}}
\newcommand{\tgft}{\textsc{TGft}}
\newcommand{\nngft}{\textsc{NnGft}}
\newcommand{\spgft}{\textsc{SpGft}}
\newcommand{\mstgft}{\textsc{MstGft}}
\newcommand{\sealshap}{\textsc{SEAL-Shap}}

\title{Gradual Fine-Tuning with Graph Routing for Multi-Source Unsupervised Domain Adaptation}

\author{Yao Ma\thanks{Correspondence to: Yao Ma \textless\texttt{yaoom@amazon.co.uk}\textgreater.} \textsuperscript{ ,}\thanks{These authors contributed equally to this work.} , Samuel Louvan\footnotemark[2] , Zhunxuan Wang\footnotemark[2] \\
Amazon \\
United Kingdom \\
\texttt{\{yaoom,slouvan,wzhunxua\}@amazon.co.uk}
}

\collasfinalcopy 

\begin{document}

\maketitle

\begin{abstract}
Multi-source unsupervised domain adaptation aims to leverage labeled data from multiple source domains for training a machine learning model to generalize well on a target domain without labels. Source domain selection plays a crucial role in determining the model's performance. It relies on the similarities amongst source and target domains. Nonetheless, existing work for source domain selection often involves heavyweight computational procedures, especially when dealing with numerous source domains and the need to identify the best ones from them. In this paper, we introduce a framework for gradual fine tuning (\gft) of machine learning models on multiple source domains. We represent multiple source domains as an undirected weighted graph. We then give a new generalization error bound for \gft~along any path within the graph, which is used to determine the optimal path corresponding to the optimal training order. With this formulation, we introduce three lightweight graph-routing strategies which tend to minimize the error bound. Our best strategy improves $2.3\%$ of accuracy over the state-of-the-art on Natural Language Inference (NLI) task and achieves competitive performance on Sentiment Analysis (SA) task, especially a $3.9\%$ improvement on a more diverse subset of data we use for SA.
\end{abstract}

\section{Introduction}\label{sec:intro}

Domain adaptation has been shown to succeed in training deep neural networks with limited data, particularly when the acquisition of labeled data can be costly in real-world applications. In practice, it is often favorable to train a model with data from related domains. Accordingly, the effectiveness of domain adaptation highly depends on the quality and similarity of the source domains' datasets. In cases where few or no labeled samples are available in the domain we target, developing a methodology to train a model without direct supervision on target domain becomes necessary. This approach is known as unsupervised domain adaptation.

Extensive research has been conducted on theoretical analysis and  empirical algorithms that minimize the generalization error (risk) of the trained model on target domain. \citet{10.5555/1795114.1795157} has demonstrated that the generalization error depends on both generalization error on the source domain and the discrepancy between source and target domains. In the pursuit of minimizing the discrepancy, certain approaches \citep{ruder-plank-2017-learning,Liu2019ReinforcedTD} have been proposed to select source domains that are close to the target domain. Nevertheless, this selection process can be costly as it introduces an additional step prior to the model training. Moreover, the source domain selection tends to discard distant domains, while we assert that the distant domains can actually provide valuable training benefits for the target domain. For this reason, we propose a lightweight and efficient gradual fine-tuning (\gft) framework that can take advantage of all available source domains. By sequentially fine-tuning a model on multiple data sources, we aim to address the limitation of existing methods and unlock the potential benefits from distant domains during the training process for the target domain. The underlying intuition of our approach is to \textit{gradually} guide a model to its optimal solution through sequentially fine-tuning the model on different source data whose distribution progressively aligns with the target domain. With this formulation, the model learns from data that spans a wide range of distributions, and ultimately leading it towards a better performance on the target domain. The gradual alignment of source data distributions with the target domain is crucial in enhancing adaptability and performance.

The contribution of our work is two-fold. First, motivated by \citet{Wang2022UnderstandingGD}, we construct theoretical analysis and give the generalization error of the proposed \gft~algorithms. Based on our theory, we introduce graph routing strategies for determining the optimal paths through an undirected graph constructed by source domains and Wasserstein distances thereof. Second, we present empirical results for the \gft~algorithms on two Natural Language Processing (NLP) tasks that are commonly used to demonstrate the effectiveness of a domain adaptation algorithms. We show that the performance of \gft~algorithms does not highly depends on the discrepancy of sources and target domains as long as there exists a path along domains such that the distance between consecutive domains is small and the data magnitude on the path is large. Both empirical and theoretical results indicate that through \gft~framework the model achieves better performance than baselines.

\section{Related Work}\label{sec:related-work}

\textbf{Domain Adaptation} aims to learn a model from source domains that generalize well to a target domain. In general, there are three types of approaches on domain adaptation, namely model based, data centric, or hybrid approaches \citep{Pan2010ASO,ramponi2020neural}. Model based approaches typically aim to learn an invariant representation for domain shift \citep{ganin2015unsupervised,DBLP:conf/icml/0002CZG19,li2020maximum}. However, there is no theoretical analysis presented in these works. Existing works \citep{NIPS2006_a2186aa7,10.5555/1795114.1795157,courty2017joint} also have proposed on determining the value of sources, including the number of samples, the quality of data, and the discrepancy between source and target.  The data centric approaches typically perform data selection to select source domains that are more similar to the target domain in terms of data distribution. Different kinds of metrics have been used to measure domain similarity, for example in NLP, Jensen Shannon similarity over word distribution \cite{ruder-plank-2017-learning} is one of the common metrics to be used. One of the downside of data selection methods is, it usually involves expensive computation in addition to the model training such as data selection using Bayesian Optimization \citep{ruder-plank-2017-learning} and Reinforcement Learning \citep{Liu2019ReinforcedTD}. Instead of performing data selection, our work attempts to eliminate this source selection stage through gradual fine-tuning.

\textbf{Gradual Domain Adaptation} (GDA) is proposed for the problem of unsupervised domain adaption which assumes the existence of a set of unsupervised datasets from intermediate domains. A pre-trained model is trained using labeled data from source domain. And then the model is trained and updated sequentially with pseudo-labeled predicted by the current model by minimizing the empirical loss w.r.t. the pseudo-labels. \cite{pmlr-v119-kumar20c} have shown the GDA achieves a small generalization error when the distribution shift between two consecutive domains is small and the error in source domain is small. \citet{Wang2022UnderstandingGD} further proved an improved generalization bound which only grows linearly with the number of intermediate domains. \citet{NEURIPS2021_45017f65} considers the problem of gradual domain adaption when the intermediate domains are not clearly defined, and proposed Intermediate Domain Labeler (IDOL) to assign scores to all unlabeled samples and group samples into different domains accordingly. 

\textbf{Learning from multiple sources.} 
As the data from a single domain could be very limited, \citet{mancini2018boosting,peng2019moment,zhao2018adversarial} consider the problem of multi-source unsupervised domain adaptation which assumes the source domain examples are multi-modal, i.e., the samples are drawn from different distributions. Multi-source domain adaptation considers the setting when training with joint data samples from multiple sources and little or no labeled training data from target domain. \citet{crammer08} presented a bound on the expected error incurred by using $\mathcal{K}$ number of data sources. By applying the bound, an optimal number of data sources to train a model can be achieved by measuring the discrepancy of data sources. \citet{NIPS2008_0e65972d} considers a similar problem which assumes the target distribution is a mixture of distributions of multiple sources. The results show that there exists a distribution weighted mixture combining rule that has a small enough loss with respect to any consistent target function and any mixture of the data distributions. Furthermore, \citet{10.5555/1795114.1795157} relaxes the assumption and shows that there exists a distribution weighted combination of the source hypotheses whose loss can be bounded with respect to the maximum loss of the source hypotheses and the Renyi divergence.

\section{Problem Setup}\label{sec:prob-setup}

In this paper, we consider a binary classification problem. We denote $\mathcal{X}$ as the feature space and $\mathcal{Y}=\{-1,+1\}$ as the label space. We assume that the feature space is compact and bounded by an $L^2$ ball, i.e., $\mathcal{X}\subseteq\{\vx\in\mathbb{R}^d:\|\vx\|_2\leq 1\}$. A domain is defined by a joint data distribution $D$ with sample space $\mathcal{X} \times \mathcal{Y}$, where $\mathcal{X}$ and $\mathcal{Y}$ are mapped by a labeling function $f:\mathcal{X}\to\mathcal{Y}$. For any domain, we assume that the data distribution is unknown, and we draw $n$ sample pairs $S=\{\vx_i,y_i\}_{i=1}^{n}$ from $D$ independently. A hypothesis (classifier) is represented as a function $h: \mathcal{X}\to\mathcal{Y}$. Let $\mathcal{H}$ denotes the hypothesis class, which is a set of classifiers. In this paper, we assume any classifier $h\in\mathcal{H}$ is Lipschitz continuous with respect to the feature vector $\vx$. More precisely, for any classifier $h\in\mathcal{H}$, there exists a real constant $R\geq0$ such that $\forall \vx, \vx^\prime\in X$, $|h(\vx)-h(\vx^\prime)|\leq R\cdot\|\vx-\vx^\prime\|_2$.

Then, for any $h$, we can define its expected loss (risk) with loss function $\mathcal{L}$ on data distribution $D$ as
\begin{equation*}
    \epsilon_D(h) = \E_{\{\vx,y\}\sim D}[\mathcal{L}(h(\vx),y)].
\end{equation*}
Similarly, we denote the empirical loss on $n$ samples as 
\begin{equation*}
    \hat\epsilon_D(h,f) = \frac{1}{n}\sum_{i=1}^n\mathcal{L}(\vx_i,y_i).
\end{equation*}
To provide bounds on the expected loss, we make an assumption that the considered loss function $\mathcal{L}$ is also Lipschitz continuous w.r.t. the input of the function.
That is, there exists a real constant $L\geq0$, such that $\forall y,y'\in \mathcal{Y}$, we have $
    |\mathcal{L}(\cdot,y)-\mathcal{L}(\cdot,y')|\leq L\|y-y'\|_2$.

In standard supervised learning, a model is trained to minimize the empirical loss (i.e., training loss) using $n$ samples. The trained model is expected to perform well on a test dataset when the test data and training data are drawn from the same distribution $D$. However, this assumption does not hold in the context of domain adaptation, where the test data (target domain) is not drawn from the same distribution as the training data (source domain(s)). 

\noindent \textbf{Problem Statement.} The objective of domain adaptation is to learn a classifier that minimizes the expected loss or risk on test data from the target domain, given the training data is from multiple source domains without knowing the underlying joint distribution. Formally, the loss is 
\begin{equation*}
    \epsilon_T(h) = \min_{h\in\mathcal{H}}\E_{\{\vx,y\}\in D_T}[\mathcal{L}(h(\vx),y)].
\end{equation*}
We assume the model has access to $D_T$ as a distribution but does not have access to its data. Instead, certain number of samples drawn from other different distributions are the only labeled resource available for training. Different from most of the single domain adaptation set-ups, we consider a problem where there are $K$ source domains and a single target domain. We denote the data distribution and data samples for the $t$-th domain with $D_t$ and $S_t$ respectively. All data distributions are unknown,  we are only given $K$ set of data samples $S_t,t=1,...,K$ drawn from distributions $D_1,...,D_K$ respectively. Each $D_t$ has different level of discrepancy with $D_T$. Existing work \citep{ramesh-kashyap-etal-2021-domain,ruder-plank-2017-learning} has been proposed to evaluate discrepancy measurement between source and target, e.g., KL-divergence \citep{kullback1951information}, Jansen-Shannon divergence \citep{Lin1991DivergenceMB}, Renyi divergence \citep{renyi1961measures}, and Wasserstein-$p$ distance \citep{Villani2009}. In this paper, we use Wasserstein-$p$ as a distance metric on a space of probability measures, which has constantly been a reliable measurement for achieving good transfer performance \citep{Villani2009, shen2018wasserstein}. For any distribution $D_1$ and $D_2$, the Wasserstein-$p$ distance is defined by the value of the following minimization problem:
\begin{equation*}
W_p(D_1,D_2)= \mathrm{inf}_{\pi\in\Gamma(D_1,D_2)}\int\|\vz_1-\vz_2\|_p\mathrm{d}\pi(\vz_1,\vz_2),
\end{equation*}
where $\Gamma(D_1,D_2)=\{\pi|\int\pi(\vz_1,\vz_2)\mathrm{d}\vz_1=D_1(\vz)$ and $\int\pi(\vz_1,\vz_2)d\vz_2=D_2(\vz), \vz\in\mathcal{X}\times \mathcal{Y}\}$ is the set of joint distributions with marginals $D_1$ and $D_2$.
In reality, the estimation of Wasserstein-$p$ distance of two distributions with finite number of samples could be challenging. The Sinkhorn algorithm \citep{chizat2020faster} provides a practical way to estimate Wasserstein distance by solving an entropy regularized minimization problem. In this work, we apply this estimator to evaluate the distance between domains.

\section{Gradual Fine-Tuning}\label{sec:gft}

In this section, we present our \gft~approach for training on multiple source domains sequentially. Our method is inspired by the fact that the generalization error of a trained classifier increases linearly with the distance between the initial and the final parameter values. Previous research \citep{10.5555/1795114.1795157} has shown that the target error depends on both the source error and the discrepancy between the source domain data distributions $D_S$ and target domain data distribution $D_T$. This relationship explains why domain adaptation often works well in practice. However, existing methods have not fully exploited these insights. 

Our \gft~approach addresses this gap by gradually updating the model based on the source domains in sequence. We use graph routing algorithms to determine the order of updates, ensuring that each update minimizes the total error on all previous source domains while maximizing the accuracy on the current one. This way we ensure that the model converges faster and performs better overall compared to traditional single-source fine-tuning approaches. The \gft~approach provides a principled framework for handling multi-source domain adaptation problems, allowing us to leverage the benefits of distant sources without sacrificing performance on the target domain.

Given $K$ labeled datasets $S_k,k=1,...,K$ from $K$ sources and an unlabeled dataset $T$ from the target domain, we want to quantify the similarity between the source domains and the target domain. To achieve this, we employ the Wasserstein-$p$ distance, specifically utilizing the Sinkhorn divergence estimation method \citep{chizat2020faster}, to
{
\parfillskip=0pt
\parskip=0pt
\par}
\begin{wrapfigure}{l}{0.5\textwidth}
    \centering
    \includegraphics[width=0.8\linewidth]{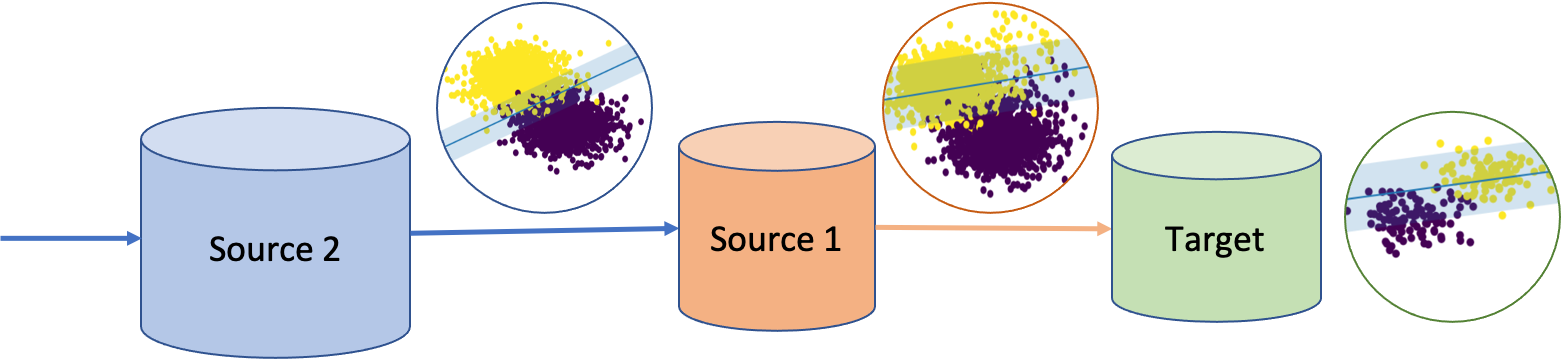}
    \caption{\gft~illustration for $2$-source domain adaptation with linear binary SVM. Source 1 has a distribution close to target but its scale is diminutive, whereas Source 2 has a large size but diverges further from the target. Model is first trained from scratch on Source 2, giving a clear hyperplane splitting two classes. Then fine-tuned on Source 1, shifting the hyperplane towards Source 1 distribution. Evaluation on target demonstrates the efficacy of \gft.}
    \label{fig:illustration}
\end{wrapfigure}
\noindent measure the distance between each pair of source domains. This yields a symmetric $(K+1)\times(K+1)$ matrix $W$ whose $i,j$-th entry is $W_p(D_i,D_j)$.

For a given disparity matrix $W$, we model those multiple domains as a Wasserstein geometric graph $G = (V, E)$, where vertices $V$ are domains and edges $E$ are pairs of domains weighted by their Wasserstein distances. Any path to the target domain in $G$ represents a \gft~trajectory to train a classifier. Notably, the resulting graph $G$ is complete when no threshold is applied to the maximum Wasserstein distance. However, to ensure that the disparity graph $G$ is meaningful and easier to interpret, we introduce a threshold value $\tau$ for the Wasserstein distance. Specifically, the disparity graph $G$ satisfies $\forall i,j\in V, (S_i, S_j)\in E$ if and only if the distance $W_{i,j}<\tau$. The threshold parameter $\tau$ plays a crucial role in ensuring the efficacy of the \gft~algorithm. Intuitively, it serves as a bound on the lengths of edges in the graph, preventing excessive errors from propagating through the network. Specifically, the threshold guarantees that every edge in the graph is bounded by a small positive constant, which is essential for maintaining a low expected error rate. By applying the threshold, the disparity graph induced by $W$ gets pruned, thereby enhancing the accuracy of the algorithm because it ensures that paths between high-discrepancy sources are not possible. On the other hand, we lack the access to target labels, indicating that direct Wasserstein distance calculation is not feasible. Therefore, we adopt a common approach in unsupervised domain adaptation, where pseudo-labels are generated by a pre-trained classifier on sources to make it Wasserstein measurable between source and target domains.

Now, we formally present our \gft~algorithm, which leverages the disparity graph $G$ defined earlier. For any path in $G$, our approach trains the model iteratively along the path using empirical risk minimization for each dataset. At each step, we start with the previously train model $\hat{h}_{t-1}$ and fine-tune it on $S_t$ by minimizing the empirical loss $\mathcal{L}_t(h)$ defined as
\begin{equation}
    \hat{h}_{t}=\frac{1}{n_t}\argmin_{h\in\mathcal{H}}\sum_{(x,y)\in S_t}\mathcal{L}(h(x), y).
\end{equation}
In summary, our \gft~algorithm provides a flexible framework to train a model on multiple datasets without merging them into a single one. While the framework itself does not explicitly define a specific path for domain adaptation, a path from the most distant source to the closest source with the minimum sum of weights may lead to better performance in unsupervised domain adaptation. An illustration of example \gft~training with two sources is shown in Figure~\ref{fig:illustration}, where the path is ``Source $2\to$ Source $1$''. We also conduct initial experiments on the 2-source example and \gft~achieves the best test accuracy compared to training on individual and combined sources. See appendix \ref{app:sim-res} for details.

\section{Theoretical Analysis}\label{sec:theoretical-analysis}

In this section, we present the generalization error bound of the classifier trained with \gft~algorithm along any path in the disparity graph $G$. We first recall the result from \citet{Wang2022UnderstandingGD} which shows the error difference of any classifier $h$ over shifted data distribution is bounded by the Wasserstein-1 distance.
\begin{lemma}
\label{lem:error-diff}
Given two joint distributions $D_1$ and $D_2$ over $X\times Y$, the expected loss of a classifier $h$ satisfies
\begin{equation}
    |\epsilon_{D_1}(h)-\epsilon_{D_2}(h)|\leq L\sqrt{R^2+1}W_1(D_1,D_2).
\end{equation}
\end{lemma}

Lemma \ref{lem:error-diff} gives the performance discrepancy bound of a classifier between two different datasets. A classifier produces similar errors on two data distributions that has smaller Wasserstein-1 distance. Applying this lemma, we bound the expected errors of two consecutive classifiers for the proposed \gft~algorithm as
\begin{equation*}
    \epsilon_{t+1}(\hat{h}_{t+1})-\epsilon_{t}(\hat{h}_{t})\leq \frac{4B\sqrt{2}L}{\sqrt{n_{t+1}}}
   +4B\sqrt{\frac{\log{1/\delta}}{2n_{t+1}}}+L\sqrt{R^2+1}W_p(D_{t+1},D_{t}),
\end{equation*}
where $\Delta_{t,t+1}=\Delta_{t+1,t}=W_p(D_{t+1},D_{t})$.The first and second terms are from generalization error bound under the assumption that the Rademacher complexity of the hypothesis space satisfies $\mathcal{R}_{n}(\mathcal{H})\leq\frac{B}{\sqrt{n}}$.
Note that although the above result is very similar to the bound in \cite{Wang2022UnderstandingGD}, this result is different since the difference between labels of two consecutive datasets is considered in this paper. This bound is essential to analyze the generalization error. The intuition is that the gradual fine-tuning works well only when the consecutive datasets are close enough. 

We then present the theorem for bounding the generalization error of the gradual fine-tuned model under stated assumptions. We follow the same line of the analysis in \cite{Wang2022UnderstandingGD} which treats the training procedure as an online learning problem. More specifically, we sequentially train the final classifier with $\kappa$ datasets consists of $\sum_{t=1}^\kappa n_t$ training samples. The result in \cite{DBLP:journals/amai/KuznetsovM20} shows the discrepancy measurement between distributions provides a tight generalization error bound for the final classifier with the target data distribution. For any $\mathbf{q_{\kappa}}=(q_1^1,...,q_1^{n_1},q_2^1...,q_2^{n_2},...,q_{\kappa}^1,...,q_{\kappa}^{n_{\kappa}})\in\mathcal{R}^{\sum_{t=1}^\kappa n_t}$, the discrepancy measurement is defined as
\begin{equation*}
    \disc(\mathbf{q_{\kappa}})=\sup_{h\in\mathcal{H}}\left(\epsilon_{\kappa}(h)-\sum_{t=1}^{\kappa}\sum_{\tau=1}^{n_t}q_t^{\tau}\epsilon_{t}(h)\right).
\end{equation*}
The following bound holds for any real number sequence $\mathbf{q}_{\kappa}$ with at least probability $1-\delta$,
\begin{equation*}
\epsilon_T(h_\kappa)\leq\sum_{t=1}^\kappa\sum_{i=1}^{n_t}q_t^i\epsilon_i(h_\kappa)+\|\mathbf{q_{\kappa}}\|_2+\disc(\mathbf{q_{\kappa}}) +6B\sqrt{4\pi\log{\sum_{t=1}^\kappa n_t}}\mathcal{R}^{\mathrm{seq}}_{n_{1:\kappa}}+B\|\mathbf{q_{\kappa}}\|_2\sqrt{8\log{1/\delta}},
\end{equation*}
where $\mathcal{R}^{\mathrm{seq}}_{n_{1:\kappa}}$ is the sequential Rademacher complexity \cite{rokhlin2017asymptotic} of $\mathcal{H}$ with loss function $\mathcal{L}$. 

\if
Given a fixed loss function $\mathcal{L}$ and hypothesis set $\mathcal{H}$, the upper bound of the discrepancy measure can be derived from Lemma \ref{lem:error-diff}.

\begin{lemma}
The discrepancy measure can be upper bounded as
\begin{equation*}
    \disc(\mathbf{q_T})\leq\rho\sqrt{R^2+1}\sum_{i=1}^{\kappa}\sum_{t=1}^{n_i}q_{i}^t\Delta_{i,T}.
\end{equation*}
\end{lemma}
\fi

By applying this result, we are able to analyze the expected error of the \gft~ algorithm with the optimal weights for discrepancy measure as $\mathbf{q_{\kappa}}=\left(\frac{1}{n_1\kappa},...\frac{1}{n_1\kappa},...,\frac{1}{n_\kappa\kappa},...,\frac{1}{n_\kappa\kappa}\right)$.Here, we present our main theory for the generalization error bound of the proposed gradual fine-tuning algorithm as follows whose proof is provided in the appendix.

\begin{theorem}
\label{thm:generalization}
The expected error of the final classifier $h_\kappa$ in the target domain $T$ is upper bounded with probability at least $1-\delta$ as
\begin{equation}\label{equ:gen-bound}
\begin{array}{lll}
    \epsilon_T(\hat{h}_\kappa)\leq&L\sqrt{R^2+1}W_p(D_T,D_\kappa)+
    \hat{\epsilon}_1(\hat{h}_1)+(1+\frac{1}{\kappa})L\sqrt{R^2+1}\sum_{t=1}^{\kappa-1}\Delta_{t,t+1}\\
    &+\frac{(4\sqrt{2}LB+2\sqrt{2}B\sqrt{\log{(1/\delta)}})(\kappa-1)}{\kappa}\sum_{t=0}^{\kappa-1}\frac{1}{\sqrt{n_{t+1}}}+
    6B\sqrt{4\pi\log{\sum_{t=1}^\kappa}n_t}\mathcal{R}_\kappa^{\mathrm{seq}}(\mathcal{H})\\
    &+\frac{B\sqrt{8\log{(1/\delta)}}+1}{\kappa}\sqrt{\sum_{t=1}^\kappa\frac{1}{n_t}}\text{.}
\end{array}
\end{equation}
\end{theorem}

Theorem~\ref{thm:generalization} indicates that \gft~achieves the minimum error bound when the model is sequentially trained along the optimal path from the furthest domain to the closest source domain with respect to the target domain. The \textbf{1-st} term in Equation \ref{equ:gen-bound} is proportional to Wasserstein distance between the last domain and the target. When all domains are distant from the target, the \textbf{1-st} term naturally becomes large. The \textbf{2-nd} term represents the path length across the selected source domains, which depends on the distance between source domains and the number of sources selected. On the other hand, sample sizes dominate the \textbf{4,5,6-th} terms. As the number of samples increases, the \textbf{4,5,6-th} term decreases, indicating that using far-away but large-size domains in learning can still lead to low generalization bound on target domain. Conversely, the opposite also holds for these terms when conditions are reversed. As long as the consecutive domains are similar enough, the generalization error remains bounded. This verify our intuition that even distant domains help learn the target domain when a good connecting path exists. Note that this generalization bound acts as a worst-case scenario for prediction errors. In practice, it guarantees that actual errors remain below this upper bound, but it doesn't necessarily mean achieving the minimum prediction error itself. We focus on the optimization of this worst-case scenario in the following.

For comparison, we also analyze the expected error bounds for two baselines (detailed analysis is in Appendix \ref{app:baseline-as} and \ref{app:baseline-cs}). First one is training with joint data from all source domains. The expected error of the trained classifier scales as the weighted Wasserstein-1 distance between each source and the target. When a domain has dominate number of samples and large enough Wasserstein distance with the target, the error on this domain will dominate the final trained classifier. The second baseline is training a model only on the closest domain. As in standard learning theorem, the risk decreases monotonically as the number of sample grows. In the case of the closest domain does not have contain enough samples, a trade-off between Wasserstein distance and the number of samples needs to be carefully considered and selected.

\section{Graph Routing}\label{sec:graph-routing}

Pivoting around the minimization of generalization error bound in Theorem~\ref{thm:generalization} for the best worst-case scenario, we present our \gft~trajectory selection strategies based on classical graph routing algorithms. As justified previously, a path in $G$ represents a \gft~trajectory. Theorem \ref{thm:generalization} thereby indicates that corresponding error bound of the \gft~trajectory can be represented by length of the path and sizes of source domain data sets on the path. Qualitatively, our objective is to route a path in $G$ that minimizes the error bound. Because the number of paths in $G$ is limited, we can exhaust every single path in $G$ to get the optimal path that minimizes the bound. However, the number of paths in $G$ grows larger than exponential functions and even factorial when $G$ is near complete \citep{jokic2022number}, which is obviously multiplied in the computational complexity of obtaining the optimal path. Therefore, in this section, we aim to explore more efficient ways to get paths with acceptable loss of optimality.

By observation, the error bound in Theorem~\ref{thm:generalization} increases w.r.t.~weights of the path $\sum_{t=1}^{\kappa-1}\Delta_{t,t+1}$, which naturally derives the first principle of path search: minimize path weights. Another more intricate factor is the magnitude, i.e. the sum of source domain data sizes $n_i$ on the path, which exists in the last three terms in Theorem \ref{thm:generalization}. For a fixed $\kappa$, all the terms decreases\footnote{For the term with sequential Rademacher complexity $\mathcal{R}^{\mathrm{seq}}_\kappa$, since $\mathcal{R}^{\mathrm{seq}}_\kappa$ has the same order as $1/\sum{n_i}$, it also decreases w.r.t.~$n_i$.} w.r.t. $n_i$. The bound decreases as adding new large sized sources into training, because $\mathcal{R}^{\mathrm{seq}}_\kappa$ mostly dominates sum of magnitude terms by its relatively large constant. We then get the second path search principle: maximize path magnitudes. Following the two principles, we obtain the optimal path $\bm{\pi}^\ast$ by formulation
\begin{equation}\label{equ:path-search}
    \bm{\pi}^\ast \approx \argmax_{\bm{\pi} \in P^\ast} \magn(\bm{\pi}),\text{ where } P^\ast = \left\{ \argmin_{\bm{\rho} \in P_G(S_i, T)} \Delta(\bm{\rho}),\, i \in [K] \right\},
\end{equation}
where $\Delta(\bm{\rho})$ is the sum of weights (i.e. Wasserstein distance) along any path $\bm{\rho}$. For each source dataset $S_i$, we first minimize the weights over all the paths from $S_i$ to $T$ in $G$, denoted by $P_G(S_i, T)$. Then we maximize magnitudes over all minimal weight paths, denoted by $P^\ast$, each of which corresponds each $S_i$ as start source. The first stage from $P_G(S_i, T)$ to $P^\ast$ is guided by graph routing. The second stage from $P^\ast$ to approximately optimal path is evaluated on the magnitude metric we defined.

\textbf{Repetitive Nearest Neighbor Search.} Our first proposed graph routing strategy employs the repetitive nearest neighbor algorithm, which selects the closest unvisited neighbor from one vertex to another until it reaches dead end. We utilize nearest neighbor algorithm with $T$ as the starting vertex and accumulate the path at each stop. Assuming we don't prune the original graph $G$ in this strategy, by going $K$ steps we exhaust all vertices and obtain an approximate length optimal path for each $S_i$ by backtracking its cumulative path until $T$:
\begin{equation*}
\begin{gathered}
    \bm{\rho}_{\mathrm{nn}}\left(S_i\right) = \left\{\left(T, f(T)\right), \left(f(T), f^2(T)\right), \dots, \left(f^{K-1}(T), S_i\right)\right\}\\
    \text{where } f(u) = \argmin_{v,\, (u, v) \in E}W_p(u, v),\, f^K(T) = S_i\text{.}
\end{gathered}
\end{equation*}
Note that nearest neighbor graph routing cannot get the exact minimal length path for each $S_i$, but it applies a greedy strategy that guarantees every next stop on the path comes from its closest non-successor, which acts as a $1$-gram safe move in \gft~sense: fine-tuning on the source closest to the next target achieves the closest behaviour to it in one-move scope. It also guarantees maximum magnitudes, as the fact that it exhausts every source domain brings a Hamilton path as one of approximate optima.

\textbf{Shortest Paths.} Classical shortest path (SP) routing gets the exact minimal length paths $P^\ast$ in Equation \ref{equ:path-search} by definition. For each source domain $S_i$, we apply Dijkstra's algorithm \citep{cormen2022introduction} to calculate the SP from $S_i$ to $T$ in $G$, i.e. the path with minimal sum of $W_p$ between every pair of adjacent domains on the path:
\begin{equation*}
    \bm{\rho}_{\mathrm{sp}}\left(S_i\right) = \left\{\left(S_i, u_1\right), \left(u_1, u_2\right), \dots, \left(u_{\kappa-1}, T\right)\right\},\,
    \text{where } \vu = \argmin_{\substack{\{u_1, \dots, u_{\kappa-1}\}\subset V\\(u_i, u_{i+1}) \in E\\u_0 = S_i,\, u_{t+1} = T}}\sum_{i = 0}^{\kappa-1} W_p(u_i, u_{i+1})\text{.}
\end{equation*}
This strategy gives exactly $P^\ast$ and guarantees to get the optimal path length in the generalization bound, but it generally doesn't generate large magnitudes because shortest paths only contain a small portion of vertices in a general graph. In this case, we define the magnitude as the sum of data size of every domain on the path. Note that we apply edge weight thresholds to $G$ for this strategy, as a compromise that enhances the magnitude of the path while making the path longer.

\textbf{Minimum Spanning Tree.} This strategy is based on minimum spanning tree (MST), which is the tree as a subgraph of $G$ that makes all vertices in $G$ connected while its edge subset has the minimal weight sum. We prune $G$ to its most lightweight connected acyclic form whose edges are short and every pair of vertices have one and only one path in between. For domain graph $G$, we apply Kruskal's algorithm \citep{cormen2022introduction} to calculate its MST. Then for each source domain $S_i$, we use the exactly one path from $S_i$ to $T$ to form the approximate $P^\ast$:
\begin{equation*}
    \bm{\rho}_{\mathrm{mst}}\left(S_i\right) \in P_{\mathrm{MST}(G)}(S_i, T), \text{ where } \forall u, v \in V,\, \abs{P_{\mathrm{MST}(G)}(u, v)} = 1 \text{.}
\end{equation*}
This strategy employs a further trade-off with the shortest path strategy between path lengths and magnitudes in the generalization bound. The Cut Property of MST states that any newly added edge to an MST forms a cycle and has the maximum weight on the cycle. As a result, MST can omit potential edges from a shortest path and take alternative routes that involve smaller edges, enhancing its magnitudes.

The proposed strategies leverage the power of graph-based representations to facilitate the seamless knowledge transfer from a source domain to a target domain while accounting for their distributional differences and magnitudes.

\section{Experimental Setup}
We evaluate our proposed \gft~methods, focusing on  sentiment analysis (SA) and Natural Language Inference (NLI) text classification tasks.

\textbf{Multi-domain Datasets.} For the SA task, we use the Amazon Review dataset \citep{blitzer-etal-2007-biographies, liu-etal-2017-adversarial}, which contains product reviews from 20 domains, annotated with binary sentiment labels (positive or negative sentiment). In this experiment, we randomly select 8 domains as shown in Table \ref{tab:overall_result_sa} for simplicity of the experiment setting. The language used in the dataset is English and Spanish. Additionally, to understand the performances of different strategies under a more difficult scenario for this task, we manually select and experiment on 4 domains out of the 8 from Amazon Review that are more diverging: \textit{books}, \textit{music}, \textit{electronics}, \textit{grocery}, which have the greatest Wassserstein-1 distance between each pair. For the NLI task, we use multi-genre Natural Language Inference \cite[\texttt{MultiNLI}]{Williams2018ABC}, which contains a sentence pair of premise and hypothesis from 5 domains. The language in the dataset is English. Each sentence pair is annotated with entailment, neutral, or contradiction labels. We binarize the label into entailment or not, by following the procedure in \cite{ma-etal-2019-domain}.

\if Table \ref{tab:dataset_statistics} shows the statistics of each dataset.

\begin{table}[t]
    \caption{Dataset Statistics for the \texttt{MultiNLI} and Amazon datasets.}
    \centering
    \begin{tabularx}{\columnwidth}{l l r r r}
        \toprule
      
         Dataset  &  Domain & \#Train & \#Dev & \#Test \\
        \midrule     
        \texttt{MultiNLI} & Fiction    & 61879 & 15471 & 1974 \\
             & Government & 61881 & 15471 & 1946\\
             & Telephone  & 66679 & 16671 & 1967\\
             & Slate      & 61845 & 15463 & 1956\\
             & Travel     & 61881 & 15471 & 1977\\
            
        \midrule
        Amazon Review     & Books & 2000 & 4787 & 5983\\
             & Software   & 2000 & 1912 & 2390\\
             & Apparel    & 2000 & 4467 & 5584\\
             & Camera     & 2000 & 2547 & 3184\\
             & Music      & 2000 & 2867 & 3584\\
             & Magazine   & 2000 & 2227 & 2784\\
             & Gourmet    & 2000 & 194  & 243\\
             & Outdoor    & 2000 & 613  & 766\\
             & Grocery    & 2000 & 830  & 1037\\
             & Electronic & 2000 & 5427 &6784\\
         \bottomrule
    \end{tabularx}
    \label{tab:dataset_statistics}
\end{table}
\fi

\textbf{Implementation \& Evaluation.} The base model for the gradual fine-tuning experiments is a BERT-based model \citep{devlin-etal-2019-bert}. 
Our gradual fine-tuning implementation is built on top of the Huggingface framework \citep{Wolf2020HuggingFacesTS}, and we use Geomloss \citep{feydy2019interpolating} to compute Wasserstein distances between domains in a dataset. For the evaluation metric, we use accuracy to compare performance between different methods on each task.  

\textbf{Baselines.} We experiment with our gradual fine-tuning (\gft) methods, namely nearest-neighbor (\nngft), shortest path (\spgft), and minimum spanning tree (\mstgft) graph routing. Additionally, we also conduct the aforementioned brute-force that exhausts every possible path in $G$ and chooses the one that minimizes the theoretical generalization bound, named \tgft. We compare our \gft~strategies to several baselines: (i) All sources 1-stage (\textsc{All Sources}): We use all source domains combined for training and evaluate it on the target domain. (ii) Closest source 1-stage (\textsc{Closest}): We use the closest source domain to the target domain and evaluate it on the target domain. We determine the closest domain by Wasserstein distance. (iii) \sealshap: A state-of-the-art method \cite{parvez-chang-2021-evaluating} that uses Shapley-based score to measure the usefulness of individual sources for transfer learning. (iv) \citet{xu2021gradual}, which gradually fine tunes on mixtures of in- and out-domain data with descending amount of out-domain data.

\begin{table*}[t]
    \sisetup{detect-weight,mode=text}
    \renewrobustcmd{\bfseries}{\fontseries{b}\selectfont}
    \renewrobustcmd{\boldmath}{}
    \newrobustcmd{\B}{\bfseries}
    \centering
    \caption{Accuracy comparison on 5 target domains from the \texttt{MultiNLI} dataset, in $\mathrm{mean}\pm\mathrm{std}$. Subscripts of average accuracies denote relative decreases to the best performance. Repeated experiments are conducted above identical set of seeds for training.}
    \begin{tabularx}{1.005\linewidth}{c c c c c c c}
        \toprule
        \multirow{2}{*}{\textbf{Method}} & \multicolumn{5}{c}{\textbf{Target Domain} } & \multirow{2}{*}{\textbf{Avg Acc.}} \\
        \cmidrule{2-6}
        & Fiction &  Government & Telephone & Slate & Travel \\
        \midrule
         \allsource & $76.62 \pm 0.67$ & $72.34 \pm 1.57$ & $71.94 \pm 1.37$ & $71.09 \pm 1.12$ & $72.47 \pm 2.02$ & $72.89_{\color{red} (\downarrow 4.7\%)}$\\
         \closest & $74.97 \pm 0.34$  & $72.88 \pm 1.19$  & $72.24 \pm 0.71$ & $73.50 \pm 1.28$ & $71.36 \pm 0.85$ & $72.99_{\color{red} (\downarrow 4.6\%)}$\\
         \midrule
         \sealshap & $74.70 \pm 1.62$  & $75.39 \pm 0.75$ & $\bm{74.63} \pm 2.05$ & $73.37 \pm 0.69$ & $75.70 \pm 3.07$ & $74.75_{\color{red} (\downarrow 2.3\%)}$ \\
         \citet{xu2021gradual} & $\bm{78.82} \pm 1.62$ & $75.29 \pm 1.11$ & $74.97 \pm 0.59$ & $75.47 \pm 1.11$ & $73.25\pm2.37$ & $75.56_{\color{red} (\downarrow 1.2\%)}$ \\
         \midrule
         \tgft & $77.43 \pm 1.78$ & $\bm{77.19} \pm 2.13$ & $72.89 \pm 2.08$ & $74.35 \pm 1.59$ & $74.68 \pm 4.43$ & $75.30_{\color{red} (\downarrow 1.6\%)}$\\
         \nngft & $78.03 \pm 2.34$ & $76.95 \pm 2.14$ & $73.74 \pm 2.19$ & $\bm{77.03} \pm 6.27$ & $\bm{76.76} \pm 2.19$ & $\bm{76.50_{\color{blue} (0.0\%)}}$\\
         \spgft & $76.40 \pm 1.31$ & $73.91 \pm 5.31$ & $73.05 \pm 1.69$ & $71.00 \pm 3.39$ & $73.14 \pm 2.85$ & $73.50_{\color{red} (\downarrow 3.9\%)}$ \\
         \mstgft & $76.18 \pm 4.39$ & $73.91 \pm 5.31$ & $73.05 \pm 1.69$ & $71.00 \pm 3.39$  & $73.14 \pm 2.85$ & $73.45_{\color{red} (\downarrow 4.0\%)}$\\
         \bottomrule
    \end{tabularx}
    \label{tab:overall_result_nli}
\end{table*}

\begin{table*}[t]
    \centering
    \caption{Accuracy comparison on 8 target domains from the multi-domain sentiment analysis dataset, in $\mathrm{mean}\pm\mathrm{std}$. Subscripts of average accuracies denote relative decreases to the best performance. Repeated experiments are conducted above identical set of seeds for training.}
    \begin{tabularx}{\linewidth}{c c c c c c c c c c}
        \toprule
        \multirow{2}{*}{\textbf{Method}} & \multicolumn{8}{c}{\textbf{Target Domain}} &  \multirow{2}{*}{\textbf{Avg Acc.}} \\
        \cmidrule{2-9}
        & Apparel & Baby & Dvd & Electronics & Beauty & Books & Grocery & Music & \\
        \midrule
         \multirow{2}{*}{\allsource}
         & $91.89$ & $91.31$ & $90.48$ & $89.96$ & $90.85$ & $90.56$ & $90.83$ & $89.96$ & $90.73_{\color{red} (\downarrow 1.1\%)}$ \\
         & $\pm 0.28$ & $\pm 0.72$ & $\pm 0.56$ & $\pm 0.85$ &
         $\pm 0.52$ & $\pm 0.45$ & $\pm 0.67$ & $\pm 0.43$ \\
         \cmidrule(r){2-10}
         \multirow{2}{*}{\closest} & $88.62$ & $88.40$ & $89.31$ & $88.34$ & $88.49$ & $88.60$ & $88.15$ & $88.82$ & $88.59_{\color{red} (\downarrow 3.4\%)}$ \\
         & $\pm 0.77$ & $\pm 0.52$ & $\pm 0.31$ & $\pm 0.31$ &
         $\pm 0.54$ & $\pm 0.36$ & $\pm 0.52$ & $\pm 0.57$ \\
         \midrule
         \multirow{2}{*}{\sealshap} & $\bm{92.13}$ & $\bm{93.13}$ & $\bm{91.00}$ & $\bm{91.68}$ & $\bm{93.90}$ & $\bm{93.23}$ & $89.43$ & $89.12$ & $\bm{91.70}_{\color{blue} (0.0\%)}$ \\
         & $\pm 0.70$ & $\pm 0.13$ & $\pm 0.13$ & $\pm 0.25$ &
         $\pm 0.32$ & $\pm 0.15$ & $\pm 0.45$ & $\pm 0.07$ \\
         \cmidrule(r){2-10}
         \multirow{2}{*}{\citet{xu2021gradual}} & $91.40$ & $92.13$ & $89.81$ & $90.08$ & $90.67$ & $89.81$ & $\bm{91.56}$ & $89.54$ & $90.63_{\color{red} (\downarrow 1.2\%)}$ \\
         & $\pm 0.11$ & $\pm 0.33$ & $\pm 0.14$ & $\pm 0.03$ &
         $\pm 0.33$ & $\pm 0.31$ & $\pm 0.62$ & $\pm 0.53$ \\
         \midrule
         \multirow{2}{*}{\tgft} & $91.93$ & $88.22$ & $90.18$ & $89.43$ & $90.51$ & $90.41$ & $89.98$ & $\bm{90.22}$ & $90.11_{\color{red} (\downarrow 1.7\%)}$ \\
         & $\pm 0.6$ & $\pm 0.54$ & $\pm 0.62$ & $\pm 0.49$ &
         $\pm 0.92$ & $\pm 0.8$ & $\pm 0.48$ & $\pm 0.26$ \\
         \cmidrule(r){2-10}
         \multirow{2}{*}{\nngft} & $91.95$ & $90.68$ & $90.31$ & $90.10$ & $90.19$ & $90.21$ & $90.37$ & $89.95$ & $90.47_{\color{red} (\downarrow 1.3\%)}$ \\
         & $\pm 0.6$ & $\pm 0.4$ & $\pm 0.8$ & $\pm 0.72$ &
         $\pm 0.53$ & $\pm 0.36$ & $\pm 0.65$ & $\pm 0.66$ \\
         \cmidrule(r){2-10}
         \multirow{2}{*}{\spgft} & $91.36$ & $88.05$ & $88.31$ & $89.50$ & $89.68$ & $88.77$ & $89.68$ & $88.52$ & $89.23_{\color{red} (\downarrow 2.7\%)}$ \\
         & $\pm 0.43$ & $\pm 0.60$ & $\pm0.64$ & $\pm 0.61$ &
         $\pm 1.02$ & $\pm 1.33$ & $\pm 0.43$ & $\pm 1.07$ \\
         \cmidrule(r){2-10}
         \multirow{2}{*}{\mstgft} & $91.35$ & $88.44$ & $87.91$ & $89.35$ & $88.98$ & $88.96$ & $89.09$ & $89.14$ & $89.15_{\color{red} (\downarrow 2.8\%)}$ \\
         & $\pm 0.42$ & $\pm 0.88$ & $\pm 0.64$ & $\pm 0.87$ & $\pm 0.94$ & 
         $\pm 0.72$ & $\pm 0.66$ & $\pm 0.98$ \\
         \bottomrule
    \end{tabularx}
    \label{tab:overall_result_sa}
\end{table*}

\begin{table*}[t]
    \centering
    \caption{Accuracy comparison on 4 distant domains from the multi-domain sentiment analysis dataset, in $\mathrm{mean}\pm\mathrm{std}$. Subscripts of average accuracies denote relative decreases to the best performance. Repeated experiments are conducted above identical set of seeds for training.}
    \begin{tabularx}{0.88\linewidth}{c c c c c c}
        \toprule
        \multirow{2}{*}{\textbf{Method}} & \multicolumn{4}{c}{\textbf{Target Domain}} &  \multirow{2}{*}{\textbf{Avg Acc.}} \\
        \cmidrule{2-5}
        & Books & Music & Electronics & Grocery\\
        \midrule
        \allsource
         & $89.71 \pm 0.31$ & $88.83 \pm 0.67$ & $\bm{87.75} \pm 0.56$ & $88.66 \pm 0.53$ & $88.73_{\color{red} (\downarrow 0.5\%)}$\\
         \closest 
         & $\bm{89.69} \pm 0.41$ & $88.98 \pm 0.22$ & $83.66 \pm 0.78$ & $88.62 \pm0.88$ & $87.73_{\color{red} (\downarrow 1.6\%)}$ \\
         \midrule
         \sealshap
         & $84.85 \pm 1.80$ & $85.91 \pm 1.52$ & $88.18 \pm 0.47$ & $84.21 \pm 1.50$ & $85.79_{\color{red} (\downarrow 3.8\%)}$ \\
        \citet{xu2021gradual}
        & $88.87 \pm 0.85$ & $88.90 \pm 0.37$ & $87.56 \pm 0.44$ & $88.72 \pm 1.52$ & $88.51_{\color{red} (\downarrow 0.7\%)}$ \\
         \midrule
         \nngft
         & $89.33 \pm 0.53$ & $\bm{89.85} \pm 0.32$ & $87.65 \pm 0.04$ & $\bm{89.85} \pm 0.82$ & $\bm{89.17}_{\color{blue} (0.0\%)}$ \\
         \bottomrule
    \end{tabularx}
    \label{tab:distant_result_sa}
\end{table*}

\section{Results \& Discussion}

\textbf{Performance on \texttt{MultiNLI}.} As shown in Table \ref{tab:overall_result_nli}, on overall average accuracy, \nngft~outperforms all the baselines including the state-of-the-art, \sealshap. On per-domain performance, \tgft~and \nngft~outperform \sealshap~on 3 and 4 target target domains, respectively. However, the results for \spgft~and \mstgft~are less positive compared to \tgft~and \nngft, this is possibly because \spgft~and \mstgft~produces shorter path consisting only 1-2 source domains, hence discarding the distant domains that can offer benefit for the performance in the target domain in the \texttt{MultiNLI} dataset. In 4 target domains, the results between \tgft~and \nngft~are the same because the produced paths are identical from both methods. The fact that \nngft~surpasses \closest~on all target domains indicates that by using only the closest source domain to the target domain is not optimal for the \texttt{MultiNLI} dataset. Additionally, the fact that \nngft~is better than \allsource~suggests that, although in terms of training data we use all source domains on both \allsource~and \nngft, gradual fine-tuning is evidently better than one-stage fine-tuning.

\textbf{Performance on SA.} \sealshap~yields the best overall results as shown in Table \ref{tab:overall_result_sa}. Per-domain wise, it outperforms all the methods in 6 out of 8 target domains. By overall average accuracy, all \gft~variants can only outperform the \closest~baseline. \tgft~obtains one better performance than \sealshap~on the Music domain. Other routing strategies, \spgft~and \mstgft, are not necessary optimal but they are still comparable to \sealshap~and other graph strategies because of their close performance to the best with low resource use of data and computation. 
From Table \ref{tab:distant_result_sa}, with the 4 distant domains, \textsc{SEAL-Shap} gets significantly hit by increased discrepancy among source domains. \nngft~outperforms all other baselines in 2 out of 4 domains and on average accuracy while stay comparable in the other 2 domains.
It is also worth noting that, \sealshap~is hugely more expensive in terms of computation. During experiments, \textsc{SEAL-Shap}~needs more than 5 hours to obtain the Shapley scores of the candidate source domains for each target domain, while \gft~only around 30 minutes in average.\footnote{See Table \ref{tab:wall-clock-time} for a more detailed wall clock time report that shows the computational efficiency of \gft~approaches.} Last but not least, our approaches do not need target labels whereas \sealshap~needs to access them in its scoring function.

\textbf{Nature of SA versus NLI.} We analyze from another aspect why \gft~performs not as significantly in the 8-domain SA experiment: SA is a relatively less complex task than NLI. In SA, detecting the sentiment polarity of some adjective keywords is empirically effective to determine the sentiment within a sentence \citep{hutto2014vader}. Such approach is common in the context of SA, where having a broader lexicon can enhance overall performance. This correlation also explains why training on all sources has strong performance in SA. On the other hand, NLI is generally considered to be a harder problem in NLP due to its requirement for more advanced linguistic comprehension and reasoning skills. NLI is essentially a textual entailment task which requires reasoning and world knowledge \citep{bowman2015large} to determine the relation between a premise and hypothesis. It is not sufficient for the NLI model to only rely on lexicon (word) encountered in a sentence, the model needs to capture the “meaning” of each sentence to then determine the relation between premise and hypothesis text fragments. We use the preceding discussion to further support \gft's capability of solving more complex tasks based on the experimental results on those tasks.

\textbf{Observation on Domain Distance.} Based on the results from both tasks, although \nngft~performs the best on the NLI task, its performance on the SA task is less effective although it's still comparable with \allsource~and \closest~baselines.  Observing the pairwise of the domain distance,  we notice that the distance between domains in the Amazon dataset is relatively smaller compared to domains in the \texttt{MultiNLI} dataset. Based on this observation, we hypothesize that \nngft~is more effective when the domains has more diverging distance. \nngft~always exhausts all source domains, but in an efficient way that it continuously seek the closest domain at each search step to try to minimize total distance travelled. As the domains in a dataset become more distant, other methods that use all/most source domains e.g. all-source, \textsc{SEAL-Shap} experience a decline in their performances. This is because growing divergence between domains cause distortion in methods that merge domains during training, and methods that choose domains in pointwise manners overlook the increasing cross-domain distance and the path length, resulting in significant prediction error bound increase, according to Theorem \ref{thm:generalization}. However, \gft~approaches follow a specific discipline that they actively controls the distance between domains which mitigates such performance regression. Experiment results reported in Table \ref{tab:distant_result_sa} also verifies this hypothesis. When we manually choose domains that are farther apart within the same dataset, the effectiveness of our proposed algorithm becomes more apparent.


\begin{table}[t]
\centering
\caption{Elapsed real time of each method on SA task, for pathfinding and training respectively. All runs are on the same computing instance with \texttt{NVIDIA Tesla V100} with cold start.}\label{tab:wall-clock-time}
\begin{tabular}{c c c c}
\toprule
\textbf{Method} & \multicolumn{1}{c}{\textbf{Pathfinding}} & \textbf{Training} & \textbf{Notes} \\ 
\midrule
\sealshap & \multicolumn{1}{c}{$< 6$ hours} & $< 10$ minutes & long pathfinding time \\
\midrule
\citet{xu2021gradual} & \multicolumn{2}{c}{$\approx 20$ minutes} & pathfinding is online in parallel with training \\
\midrule
All \gft~variants & \multicolumn{1}{c}{$\approx 30$ minutes} & $<10$ minutes & \makecell{only $\approx 5$ minutes pathfinding \\ without pseudo label generation} \\
\bottomrule
\end{tabular}
\end{table}

\begin{figure}[t]
    \centering
    \subfigure[\texttt{MultiNLI}]{
    \includegraphics[width=0.45\textwidth]{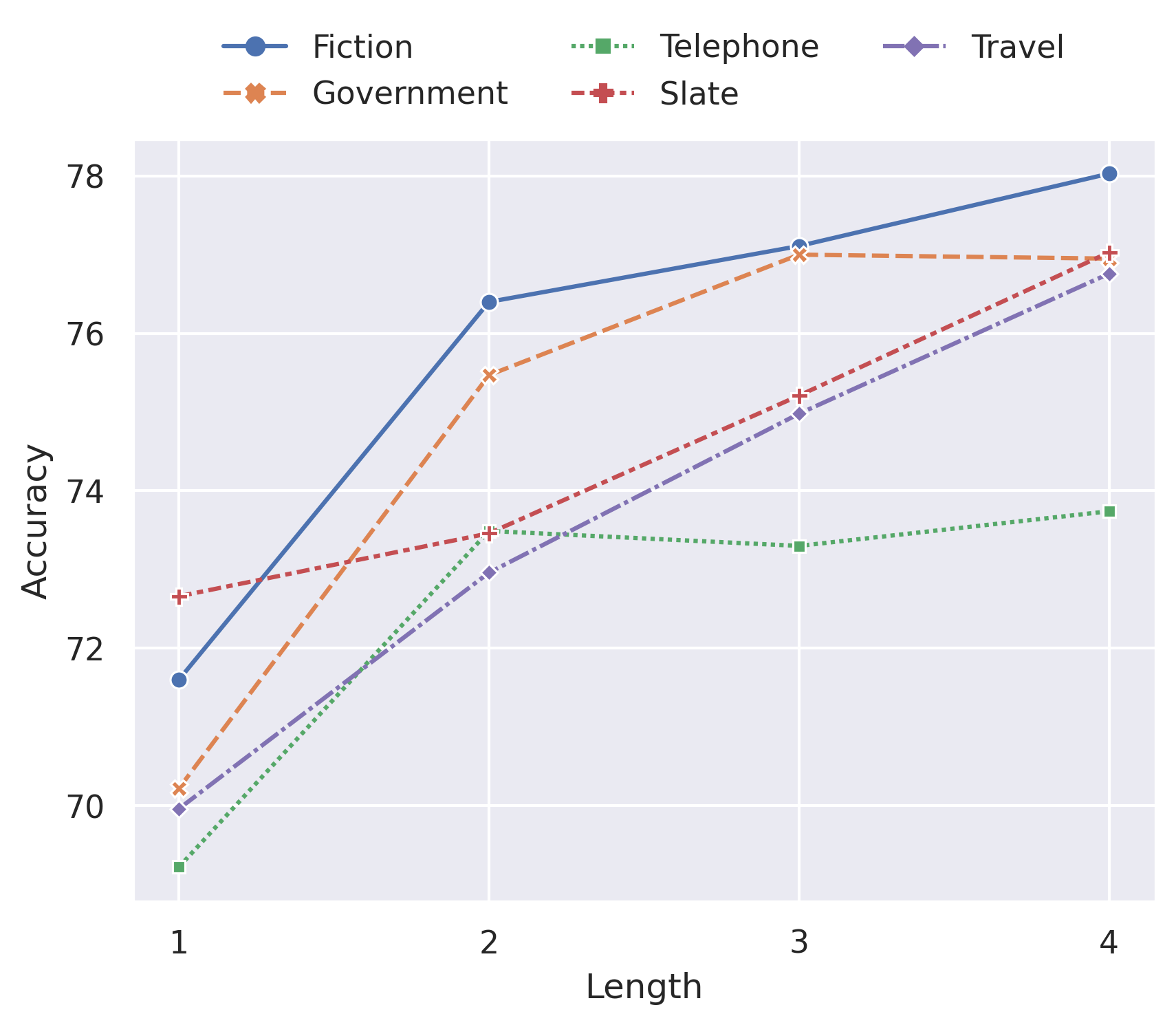}
    }\hfill
    \subfigure[Sentiment Analysis]{
    \includegraphics[width=0.45\textwidth]{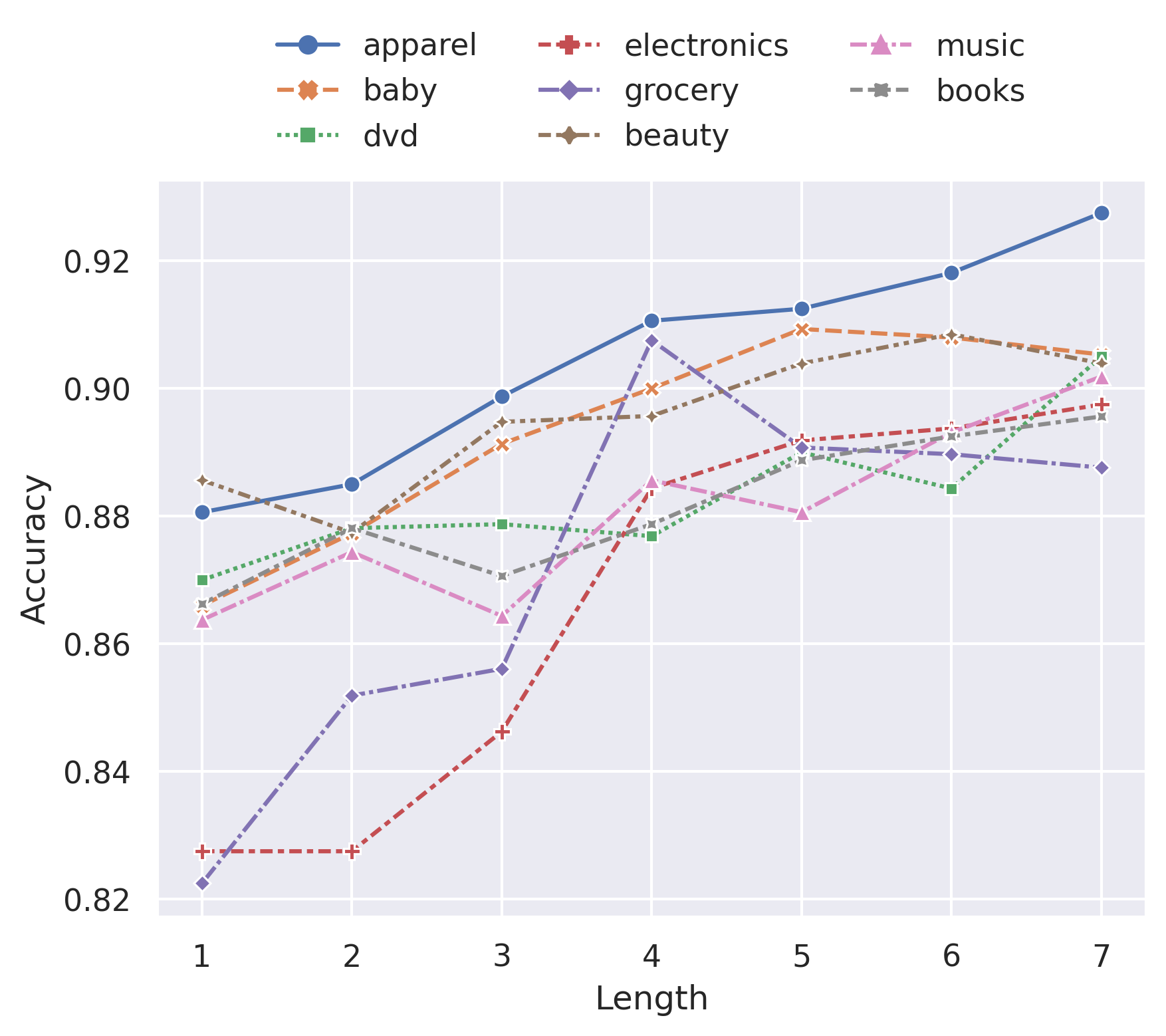}
    }
    \caption{Accuracy ablation on different path length on two datasets. The $x$-axis indicates the path length of \nngft, i.e.~number of source domains that is included in \nngft. The $y$-axis indicates the accuracy on a particular path length.}
    \label{fig:different_path_length}
\end{figure}

\textbf{Ablation Study on Path Length.} We also perform an ablation study by experimenting with different path lengths of \gft~algorithms. Given a particular sequence of fine-tuning of a target domain produced by \gft~we try to examine the behaviour of the model when we exclude a number of source domains from the sequence. For example, if the \gft~sequence of \textit{travel} target domain in \texttt{MultiNLI} is: \textit{slate} $\rightarrow$ \textit{telephone} $\rightarrow$ \textit{government} $\rightarrow$ \textit{fiction} (path length of 4). To produce path length of 3, we simply remove the furthest source domain, \textit{slate}, and measure the performance. We repeat the same step for different path length. Figure \ref{fig:different_path_length} shows the accuracy for \nngft~with different path lengths on \texttt{MultiNLI} and SA. We observe that as we include more source domains in \nngft, the trend of accuracy on most target domains are also increasing for both datasets.

\textbf{Notes on \tgft.} In the two experiments conducted, we notice that \tgft~generally performs less effectively than \nngft. It is true that generalization bound value for \tgft~is theoretically the lowest among all \gft~approaches, as it explores all possible paths and selects the one that minimizes the bound. However, as discussed above, \tgft~only guarantees the \textit{minimax} prediction error, but not the \textit{minimum} prediction error. This observation indicates that optimizing the worst-case scenario does not always yield the most practical results in \gft. It inspires deeper exploration into the factors influencing what's under the upper bound, e.g.~its distribution, to approach more optimal pathfinding strategies.

\textbf{Notes on \spgft~and \textsc{MstGft}.} For the other two graph routing strategies, i.e.~\spgft~and \textsc{MstGft}, we also highlight that they make significant trade-offs between resource consumption which reflects training computational cost, and slight performance reduction, regardless of low pathfinding cost already. Unlike \nngft~which exhaust all the domains on the fine tuning path, \spgft~and \textsc{MstGft} only select small portion of the domains and follow short paths that link those domains together, indicating low resource consumption and faster training, though they lead to a slight performance decrease. All \gft~strategies work significantly fast in pathfinding, while remaining the flexibility between training costs, as determined by the identified paths connecting selected source domains, and model performance.


\section{Conclusion}

We conduct theoretical and experimental studies on gradual fine-tuning (\gft) in multi-source unsupervised domain adaptation setting. We show that theoretically, using all source domains through \gft~minimizes the generalization error. Our experiment results show that even \textit{without} source domains selection, the adapted model from \gft~outperforms state-of-the-art method in Natural Language Inference (NLI) task and achieve comparable performance in the sentiment analysis (SA) task. We observe that (i) \gft\ is more effective when the Wasserstein distance between source domains and target are more diverge. Including distant source domain through gradual fine-tuning can improve the adapted model on the intermediate domains which is beneficial for the final target domain eventually. (ii) Path optimality for \gft\ is still an open question as our graph routing strategies are focused on mitigating worst-case scenario, and are only close to but not strictly optimal. (iii) Current graph routing strategies can hardly scale large graphs constructed by too many source domains because of computational complexity quadratic increase. We believe that our findings can be applied to more complex NLP tasks in the context of multi-source domain adaptation.

\subsubsection*{Acknowledgments}
We would like to thank Gerrit van den Burg, Jean Baptiste Faddoul, Pavel Tyletski, Nithish Kannen, Wei Liu and the anonymous reviewers for their valuable and constructive feedback. We also thank Murat Sensoy and Abhishek Tripathi for directional advice.

\bibliography{collas2024_conference}

\begin{thebibliography}{40}
\providecommand{\natexlab}[1]{#1}
\providecommand{\url}[1]{\texttt{#1}}
\expandafter\ifx\csname urlstyle\endcsname\relax
  \providecommand{\doi}[1]{doi: #1}\else
  \providecommand{\doi}{doi: \begingroup \urlstyle{rm}\Url}\fi

\bibitem[Blitzer et~al.(2007)Blitzer, Dredze, and
  Pereira]{blitzer-etal-2007-biographies}
John Blitzer, Mark Dredze, and Fernando Pereira.
\newblock Biographies, {B}ollywood, boom-boxes and blenders: Domain adaptation
  for sentiment classification.
\newblock In \emph{Proceedings of the 45th Annual Meeting of the Association of
  Computational Linguistics}, pp.\  440--447, Prague, Czech Republic, June
  2007. Association for Computational Linguistics.
\newblock URL \url{https://aclanthology.org/P07-1056}.

\bibitem[Bowman et~al.(2015)Bowman, Angeli, Potts, and
  Manning]{bowman2015large}
Samuel~R. Bowman, Gabor Angeli, Christopher Potts, and Christopher~D. Manning.
\newblock A large annotated corpus for learning natural language inference.
\newblock In Llu{\'\i}s M{\`a}rquez, Chris Callison-Burch, and Jian Su (eds.),
  \emph{Proceedings of the 2015 Conference on Empirical Methods in Natural
  Language Processing}, pp.\  632--642, Lisbon, Portugal, September 2015.
  Association for Computational Linguistics.
\newblock \doi{10.18653/v1/D15-1075}.
\newblock URL \url{https://aclanthology.org/D15-1075}.

\bibitem[Chen \& Chao(2021)Chen and Chao]{NEURIPS2021_45017f65}
Hong-You Chen and Wei-Lun Chao.
\newblock Gradual domain adaptation without indexed intermediate domains.
\newblock In M.~Ranzato, A.~Beygelzimer, Y.~Dauphin, P.S. Liang, and J.~Wortman
  Vaughan (eds.), \emph{Advances in Neural Information Processing Systems},
  volume~34, pp.\  8201--8214. Curran Associates, Inc., 2021.
\newblock URL
  \url{https://proceedings.neurips.cc/paper_files/paper/2021/file/45017f6511f91be700fda3d118034994-Paper.pdf}.

\bibitem[Chizat et~al.(2020)Chizat, Roussillon, L\'{e}ger, Vialard, and
  Peyr\'{e}]{chizat2020faster}
L\'{e}na\"{\i}c Chizat, Pierre Roussillon, Flavien L\'{e}ger,
  Fran\c{c}ois-Xavier Vialard, and Gabriel Peyr\'{e}.
\newblock Faster wasserstein distance estimation with the sinkhorn divergence.
\newblock In H.~Larochelle, M.~Ranzato, R.~Hadsell, M.F. Balcan, and H.~Lin
  (eds.), \emph{Advances in Neural Information Processing Systems}, volume~33,
  pp.\  2257--2269. Curran Associates, Inc., 2020.
\newblock URL
  \url{https://proceedings.neurips.cc/paper_files/paper/2020/file/17f98ddf040204eda0af36a108cbdea4-Paper.pdf}.

\bibitem[Cormen et~al.(2022)Cormen, Leiserson, Rivest, and
  Stein]{cormen2022introduction}
Thomas~H Cormen, Charles~E Leiserson, Ronald~L Rivest, and Clifford Stein.
\newblock \emph{Introduction to algorithms}.
\newblock MIT press, 2022.

\bibitem[Courty et~al.(2017)Courty, Flamary, Habrard, and
  Rakotomamonjy]{courty2017joint}
Nicolas Courty, R\'{e}mi Flamary, Amaury Habrard, and Alain Rakotomamonjy.
\newblock Joint distribution optimal transportation for domain adaptation.
\newblock In I.~Guyon, U.~Von Luxburg, S.~Bengio, H.~Wallach, R.~Fergus,
  S.~Vishwanathan, and R.~Garnett (eds.), \emph{Advances in Neural Information
  Processing Systems}, volume~30. Curran Associates, Inc., 2017.
\newblock URL
  \url{https://proceedings.neurips.cc/paper_files/paper/2017/file/0070d23b06b1486a538c0eaa45dd167a-Paper.pdf}.

\bibitem[Crammer et~al.(2008)Crammer, Kearns, and Wortman]{crammer08}
Koby Crammer, Michael Kearns, and Jennifer Wortman.
\newblock Learning from multiple sources.
\newblock \emph{Journal of Machine Learning Research}, 9\penalty0
  (57):\penalty0 1757--1774, 2008.
\newblock URL \url{http://jmlr.org/papers/v9/crammer08a.html}.

\bibitem[Devlin et~al.(2019)Devlin, Chang, Lee, and
  Toutanova]{devlin-etal-2019-bert}
Jacob Devlin, Ming-Wei Chang, Kenton Lee, and Kristina Toutanova.
\newblock {BERT}: Pre-training of deep bidirectional transformers for language
  understanding.
\newblock In \emph{Proceedings of the 2019 Conference of the North {A}merican
  Chapter of the Association for Computational Linguistics: Human Language
  Technologies, Volume 1 (Long and Short Papers)}, pp.\  4171--4186,
  Minneapolis, Minnesota, June 2019. Association for Computational Linguistics.
\newblock \doi{10.18653/v1/N19-1423}.
\newblock URL \url{https://aclanthology.org/N19-1423}.

\bibitem[Feydy et~al.(2019)Feydy, S\'{e}journ\'{e}, Vialard, Amari, Trouve, and
  Peyr\'{e}]{feydy2019interpolating}
Jean Feydy, Thibault S\'{e}journ\'{e}, Fran\c{c}ois-Xavier Vialard, Shun-ichi
  Amari, Alain Trouve, and Gabriel Peyr\'{e}.
\newblock Interpolating between optimal transport and mmd using sinkhorn
  divergences.
\newblock In Kamalika Chaudhuri and Masashi Sugiyama (eds.), \emph{Proceedings
  of the Twenty-Second International Conference on Artificial Intelligence and
  Statistics}, volume~89 of \emph{Proceedings of Machine Learning Research},
  pp.\  2681--2690. PMLR, 16--18 Apr 2019.
\newblock URL \url{https://proceedings.mlr.press/v89/feydy19a.html}.

\bibitem[Ganin \& Lempitsky(2015)Ganin and Lempitsky]{ganin2015unsupervised}
Yaroslav Ganin and Victor Lempitsky.
\newblock Unsupervised domain adaptation by backpropagation.
\newblock In Francis Bach and David Blei (eds.), \emph{Proceedings of the 32nd
  International Conference on Machine Learning}, volume~37 of \emph{Proceedings
  of Machine Learning Research}, pp.\  1180--1189, Lille, France, 07--09 Jul
  2015. PMLR.
\newblock URL \url{https://proceedings.mlr.press/v37/ganin15.html}.

\bibitem[Huang et~al.(2006)Huang, Gretton, Borgwardt, Sch\"{o}lkopf, and
  Smola]{NIPS2006_a2186aa7}
Jiayuan Huang, Arthur Gretton, Karsten Borgwardt, Bernhard Sch\"{o}lkopf, and
  Alex Smola.
\newblock Correcting sample selection bias by unlabeled data.
\newblock In B.~Sch\"{o}lkopf, J.~Platt, and T.~Hoffman (eds.), \emph{Advances
  in Neural Information Processing Systems}, volume~19. MIT Press, 2006.
\newblock URL
  \url{https://proceedings.neurips.cc/paper/2006/file/a2186aa7c086b46ad4e8bf81e2a3a19b-Paper.pdf}.

\bibitem[Hutto \& Gilbert(2014)Hutto and Gilbert]{hutto2014vader}
Clayton Hutto and Eric Gilbert.
\newblock Vader: A parsimonious rule-based model for sentiment analysis of
  social media text.
\newblock \emph{Proceedings of the International AAAI Conference on Web and
  Social Media}, 8\penalty0 (1):\penalty0 216--225, May 2014.
\newblock \doi{10.1609/icwsm.v8i1.14550}.
\newblock URL \url{https://ojs.aaai.org/index.php/ICWSM/article/view/14550}.

\bibitem[Joki{\'c} \& Van~Mieghem(2022)Joki{\'c} and
  Van~Mieghem]{jokic2022number}
Ivan Joki{\'c} and Piet Van~Mieghem.
\newblock Number of paths in a graph.
\newblock \emph{arXiv preprint arXiv:2209.08840}, 2022.

\bibitem[Kullback \& Leibler(1951)Kullback and
  Leibler]{kullback1951information}
S.~Kullback and R.~A. Leibler.
\newblock {On Information and Sufficiency}.
\newblock \emph{The Annals of Mathematical Statistics}, 22\penalty0
  (1):\penalty0 79 -- 86, 1951.
\newblock \doi{10.1214/aoms/1177729694}.
\newblock URL \url{https://doi.org/10.1214/aoms/1177729694}.

\bibitem[Kumar et~al.(2020)Kumar, Ma, and Liang]{pmlr-v119-kumar20c}
Ananya Kumar, Tengyu Ma, and Percy Liang.
\newblock Understanding self-training for gradual domain adaptation.
\newblock In Hal~Daumé III and Aarti Singh (eds.), \emph{Proceedings of the
  37th International Conference on Machine Learning}, volume 119 of
  \emph{Proceedings of Machine Learning Research}, pp.\  5468--5479. PMLR,
  13--18 Jul 2020.
\newblock URL \url{https://proceedings.mlr.press/v119/kumar20c.html}.

\bibitem[Kuznetsov \& Mohri(2020)Kuznetsov and
  Mohri]{DBLP:journals/amai/KuznetsovM20}
Vitaly Kuznetsov and Mehryar Mohri.
\newblock Discrepancy-based theory and algorithms for forecasting
  non-stationary time series.
\newblock \emph{Ann. Math. Artif. Intell.}, 88\penalty0 (4):\penalty0 367--399,
  2020.
\newblock \doi{10.1007/s10472-019-09683-1}.
\newblock URL \url{https://doi.org/10.1007/s10472-019-09683-1}.

\bibitem[Li et~al.(2021)Li, Chen, Ding, Zhu, Lu, and Shen]{li2020maximum}
Jingjing Li, Erpeng Chen, Zhengming Ding, Lei Zhu, Ke~Lu, and Heng~Tao Shen.
\newblock Maximum density divergence for domain adaptation.
\newblock \emph{IEEE Transactions on Pattern Analysis and Machine
  Intelligence}, 43\penalty0 (11):\penalty0 3918--3930, 2021.
\newblock \doi{10.1109/TPAMI.2020.2991050}.

\bibitem[Lin(1991)]{Lin1991DivergenceMB}
Jianhua Lin.
\newblock Divergence measures based on the shannon entropy.
\newblock \emph{IEEE Transactions on Information Theory}, 37\penalty0
  (1):\penalty0 145--151, 1991.
\newblock \doi{10.1109/18.61115}.

\bibitem[Liu et~al.(2019)Liu, Song, Zou, and Zhang]{Liu2019ReinforcedTD}
Miaofeng Liu, Yan Song, Hongbin Zou, and Tong Zhang.
\newblock Reinforced training data selection for domain adaptation.
\newblock In Anna Korhonen, David Traum, and Llu{\'\i}s M{\`a}rquez (eds.),
  \emph{Proceedings of the 57th Annual Meeting of the Association for
  Computational Linguistics}, pp.\  1957--1968, Florence, Italy, July 2019.
  Association for Computational Linguistics.
\newblock \doi{10.18653/v1/P19-1189}.
\newblock URL \url{https://aclanthology.org/P19-1189}.

\bibitem[Liu et~al.(2017)Liu, Qiu, and Huang]{liu-etal-2017-adversarial}
Pengfei Liu, Xipeng Qiu, and Xuanjing Huang.
\newblock Adversarial multi-task learning for text classification.
\newblock In \emph{Proceedings of the 55th Annual Meeting of the Association
  for Computational Linguistics (Volume 1: Long Papers)}, pp.\  1--10,
  Vancouver, Canada, July 2017. Association for Computational Linguistics.
\newblock \doi{10.18653/v1/P17-1001}.
\newblock URL \url{https://aclanthology.org/P17-1001}.

\bibitem[Ma et~al.(2019)Ma, Xu, Wang, Nallapati, and
  Xiang]{ma-etal-2019-domain}
Xiaofei Ma, Peng Xu, Zhiguo Wang, Ramesh Nallapati, and Bing Xiang.
\newblock Domain adaptation with {BERT}-based domain classification and data
  selection.
\newblock In \emph{Proceedings of the 2nd Workshop on Deep Learning Approaches
  for Low-Resource NLP (DeepLo 2019)}, pp.\  76--83, Hong Kong, China, November
  2019. Association for Computational Linguistics.
\newblock \doi{10.18653/v1/D19-6109}.
\newblock URL \url{https://aclanthology.org/D19-6109}.

\bibitem[Mancini et~al.(2018)Mancini, Porzi, Bulò, Caputo, and
  Ricci]{mancini2018boosting}
Massimiliano Mancini, Lorenzo Porzi, Samuel~Rota Bulò, Barbara Caputo, and
  Elisa Ricci.
\newblock Boosting domain adaptation by discovering latent domains.
\newblock In \emph{2018 IEEE/CVF Conference on Computer Vision and Pattern
  Recognition}, pp.\  3771--3780, 2018.
\newblock \doi{10.1109/CVPR.2018.00397}.

\bibitem[Mansour et~al.(2008)Mansour, Mohri, and
  Rostamizadeh]{NIPS2008_0e65972d}
Yishay Mansour, Mehryar Mohri, and Afshin Rostamizadeh.
\newblock Domain adaptation with multiple sources.
\newblock In D.~Koller, D.~Schuurmans, Y.~Bengio, and L.~Bottou (eds.),
  \emph{Advances in Neural Information Processing Systems}, volume~21. Curran
  Associates, Inc., 2008.
\newblock URL
  \url{https://proceedings.neurips.cc/paper/2008/file/0e65972dce68dad4d52d063967f0a705-Paper.pdf}.

\bibitem[Mansour et~al.(2009)Mansour, Mohri, and
  Rostamizadeh]{10.5555/1795114.1795157}
Yishay Mansour, Mehryar Mohri, and Afshin Rostamizadeh.
\newblock Multiple source adaptation and the r\'{e}nyi divergence.
\newblock In \emph{Proceedings of the Twenty-Fifth Conference on Uncertainty in
  Artificial Intelligence}, UAI '09, pp.\  367–374, Arlington, Virginia, USA,
  2009. AUAI Press.
\newblock ISBN 9780974903958.

\bibitem[Pan \& Yang(2010)Pan and Yang]{Pan2010ASO}
Sinno~Jialin Pan and Qiang Yang.
\newblock A survey on transfer learning.
\newblock \emph{IEEE Transactions on Knowledge and Data Engineering},
  22\penalty0 (10):\penalty0 1345--1359, 2010.
\newblock \doi{10.1109/TKDE.2009.191}.

\bibitem[Parvez \& Chang(2021)Parvez and Chang]{parvez-chang-2021-evaluating}
Md~Rizwan Parvez and Kai-Wei Chang.
\newblock Evaluating the values of sources in transfer learning.
\newblock In \emph{Proceedings of the 2021 Conference of the North American
  Chapter of the Association for Computational Linguistics: Human Language
  Technologies}, pp.\  5084--5116, Online, June 2021. Association for
  Computational Linguistics.
\newblock \doi{10.18653/v1/2021.naacl-main.402}.
\newblock URL \url{https://aclanthology.org/2021.naacl-main.402}.

\bibitem[Peng et~al.(2019)Peng, Bai, Xia, Huang, Saenko, and
  Wang]{peng2019moment}
Xingchao Peng, Qinxun Bai, Xide Xia, Zijun Huang, Kate Saenko, and Bo~Wang.
\newblock Moment matching for multi-source domain adaptation.
\newblock In \emph{2019 IEEE/CVF International Conference on Computer Vision
  (ICCV)}, pp.\  1406--1415, 2019.
\newblock \doi{10.1109/ICCV.2019.00149}.

\bibitem[Ramesh~Kashyap et~al.(2021)Ramesh~Kashyap, Hazarika, Kan, and
  Zimmermann]{ramesh-kashyap-etal-2021-domain}
Abhinav Ramesh~Kashyap, Devamanyu Hazarika, Min-Yen Kan, and Roger Zimmermann.
\newblock Domain divergences: A survey and empirical analysis.
\newblock In \emph{Proceedings of the 2021 Conference of the North American
  Chapter of the Association for Computational Linguistics: Human Language
  Technologies}, pp.\  1830--1849, Online, June 2021. Association for
  Computational Linguistics.
\newblock \doi{10.18653/v1/2021.naacl-main.147}.
\newblock URL \url{https://aclanthology.org/2021.naacl-main.147}.

\bibitem[Ramponi \& Plank(2020)Ramponi and Plank]{ramponi2020neural}
Alan Ramponi and Barbara Plank.
\newblock Neural unsupervised domain adaptation in nlp—a survey.
\newblock In \emph{Proceedings of the 28th International Conference on
  Computational Linguistics}, pp.\  6838--6855, 2020.

\bibitem[R{\'e}nyi(1961)]{renyi1961measures}
Alfr{\'e}d R{\'e}nyi.
\newblock On measures of entropy and information.
\newblock In \emph{Proceedings of the fourth Berkeley symposium on mathematical
  statistics and probability, volume 1: contributions to the theory of
  statistics}, volume~4, pp.\  547--562. University of California Press, 1961.

\bibitem[Rokhlin(2017)]{rokhlin2017asymptotic}
Dmitry~B Rokhlin.
\newblock Asymptotic sequential rademacher complexity of a finite function
  class.
\newblock \emph{Archiv der Mathematik}, 108:\penalty0 325--335, 2017.

\bibitem[Ruder \& Plank(2017)Ruder and Plank]{ruder-plank-2017-learning}
Sebastian Ruder and Barbara Plank.
\newblock Learning to select data for transfer learning with {B}ayesian
  optimization.
\newblock In \emph{Proceedings of the 2017 Conference on Empirical Methods in
  Natural Language Processing}, pp.\  372--382, Copenhagen, Denmark, September
  2017. Association for Computational Linguistics.
\newblock \doi{10.18653/v1/D17-1038}.
\newblock URL \url{https://aclanthology.org/D17-1038}.

\bibitem[Shen et~al.(2018)Shen, Qu, Zhang, and Yu]{shen2018wasserstein}
Jian Shen, Yanru Qu, Weinan Zhang, and Yong Yu.
\newblock Wasserstein distance guided representation learning for domain
  adaptation.
\newblock In \emph{Proceedings of the Thirty-Second AAAI Conference on
  Artificial Intelligence and Thirtieth Innovative Applications of Artificial
  Intelligence Conference and Eighth AAAI Symposium on Educational Advances in
  Artificial Intelligence}, AAAI'18/IAAI'18/EAAI'18. AAAI Press, 2018.
\newblock ISBN 978-1-57735-800-8.

\bibitem[Villani(2009)]{Villani2009}
C{\'e}dric Villani.
\newblock \emph{The Wasserstein distances}, pp.\  93--111.
\newblock Springer Berlin Heidelberg, Berlin, Heidelberg, 2009.
\newblock ISBN 978-3-540-71050-9.
\newblock \doi{10.1007/978-3-540-71050-9_6}.
\newblock URL \url{https://doi.org/10.1007/978-3-540-71050-9_6}.

\bibitem[Wang et~al.(2022)Wang, Li, and Zhao]{Wang2022UnderstandingGD}
Haoxiang Wang, Bo~Li, and Han Zhao.
\newblock Understanding gradual domain adaptation: Improved analysis, optimal
  path and beyond.
\newblock In Kamalika Chaudhuri, Stefanie Jegelka, Le~Song, Csaba Szepesvari,
  Gang Niu, and Sivan Sabato (eds.), \emph{Proceedings of the 39th
  International Conference on Machine Learning}, volume 162 of
  \emph{Proceedings of Machine Learning Research}, pp.\  22784--22801. PMLR,
  17--23 Jul 2022.
\newblock URL \url{https://proceedings.mlr.press/v162/wang22n.html}.

\bibitem[Williams et~al.(2018)Williams, Nangia, and Bowman]{Williams2018ABC}
Adina Williams, Nikita Nangia, and Samuel Bowman.
\newblock A broad-coverage challenge corpus for sentence understanding through
  inference.
\newblock In Marilyn Walker, Heng Ji, and Amanda Stent (eds.),
  \emph{Proceedings of the 2018 Conference of the North {A}merican Chapter of
  the Association for Computational Linguistics: Human Language Technologies,
  Volume 1 (Long Papers)}, pp.\  1112--1122, New Orleans, Louisiana, June 2018.
  Association for Computational Linguistics.
\newblock \doi{10.18653/v1/N18-1101}.
\newblock URL \url{https://aclanthology.org/N18-1101}.

\bibitem[Wolf et~al.(2020)Wolf, Debut, Sanh, Chaumond, Delangue, Moi, Cistac,
  Rault, Louf, Funtowicz, Davison, Shleifer, von Platen, Ma, Jernite, Plu, Xu,
  Le~Scao, Gugger, Drame, Lhoest, and Rush]{Wolf2020HuggingFacesTS}
Thomas Wolf, Lysandre Debut, Victor Sanh, Julien Chaumond, Clement Delangue,
  Anthony Moi, Pierric Cistac, Tim Rault, Remi Louf, Morgan Funtowicz, Joe
  Davison, Sam Shleifer, Patrick von Platen, Clara Ma, Yacine Jernite, Julien
  Plu, Canwen Xu, Teven Le~Scao, Sylvain Gugger, Mariama Drame, Quentin Lhoest,
  and Alexander Rush.
\newblock Transformers: State-of-the-art natural language processing.
\newblock In Qun Liu and David Schlangen (eds.), \emph{Proceedings of the 2020
  Conference on Empirical Methods in Natural Language Processing: System
  Demonstrations}, pp.\  38--45, Online, October 2020. Association for
  Computational Linguistics.
\newblock \doi{10.18653/v1/2020.emnlp-demos.6}.
\newblock URL \url{https://aclanthology.org/2020.emnlp-demos.6}.

\bibitem[Xu et~al.(2021)Xu, Ebner, Yarmohammadi, White, Van~Durme, and
  Murray]{xu2021gradual}
Haoran Xu, Seth Ebner, Mahsa Yarmohammadi, Aaron~Steven White, Benjamin
  Van~Durme, and Kenton Murray.
\newblock Gradual fine-tuning for low-resource domain adaptation.
\newblock In Eyal Ben-David, Shay Cohen, Ryan McDonald, Barbara Plank, Roi
  Reichart, Guy Rotman, and Yftah Ziser (eds.), \emph{Proceedings of the Second
  Workshop on Domain Adaptation for NLP}, pp.\  214--221, Kyiv, Ukraine, April
  2021. Association for Computational Linguistics.
\newblock URL \url{https://aclanthology.org/2021.adaptnlp-1.22}.

\bibitem[Zhao et~al.(2018)Zhao, Zhang, Wu, Moura, Costeira, and
  Gordon]{zhao2018adversarial}
Han Zhao, Shanghang Zhang, Guanhang Wu, Jos\'{e} M.~F. Moura, Joao~P Costeira,
  and Geoffrey~J Gordon.
\newblock Adversarial multiple source domain adaptation.
\newblock In S.~Bengio, H.~Wallach, H.~Larochelle, K.~Grauman, N.~Cesa-Bianchi,
  and R.~Garnett (eds.), \emph{Advances in Neural Information Processing
  Systems}, volume~31. Curran Associates, Inc., 2018.
\newblock URL
  \url{https://proceedings.neurips.cc/paper_files/paper/2018/file/717d8b3d60d9eea997b35b02b6a4e867-Paper.pdf}.

\bibitem[Zhao et~al.(2019)Zhao, des Combes, Zhang, and
  Gordon]{DBLP:conf/icml/0002CZG19}
Han Zhao, Remi~Tachet des Combes, Kun Zhang, and Geoffrey~J. Gordon.
\newblock On learning invariant representations for domain adaptation.
\newblock In Kamalika Chaudhuri and Ruslan Salakhutdinov (eds.),
  \emph{Proceedings of the 36th International Conference on Machine Learning,
  {ICML} 2019, 9-15 June 2019, Long Beach, California, {USA}}, volume~97 of
  \emph{Proceedings of Machine Learning Research}, pp.\  7523--7532. {PMLR},
  2019.
\newblock URL \url{http://proceedings.mlr.press/v97/zhao19a.html}.

\end{thebibliography}
\bibliographystyle{collas2024_conference}

\newpage
\appendix

\section{Proofs for Theoretical Analysis}

Let's first recall the following general theorem.

\begin{lemma}
Let $D$ be a joint distribution over $\mathcal{X} \times \mathcal{Y}$ and $l$ be a $B$-bounded loss function that is $L$-Lipschitz in the $2$-norm in the first argument. For a given function space $\mathcal{F}$ and $f\in \mathcal{F}$, let $f^\ast=\arg\min_{f\in \mathcal{F}}\mathrm{E}_{C}[l(x,y)]$ and $\hat{f}=\argmin_{f\in \mathcal{F}}\frac{1}{n}\sum_{i=1}^{n}l(f(x_i),y_i)$ be the empirical and population loss minimizers. Then for any $\delta>0$, with probability at least $1-\delta$,
\begin{equation*}
    l(\hat{f})-l(f^\ast)\leq 2\sqrt{2}L\mathcal{R}(\mathcal{F})+2B\sqrt{\frac{\log(1/\delta)}{2m}}.
\end{equation*}
\end{lemma}

The next lemma shows the error difference over shifted domains.
\begin{lemma}\label{lem:loss-diff}
We further assume that the classifier in $\mathcal{F}$ is $R$-Lipschitz continuous, then 
\begin{equation*}
    |l(f,D_1)-l(f,D_2)|\leq L\sqrt{R^2+1}W_p(D_1,D_2).
\end{equation*}
\end{lemma}

Lemma \ref{lem:loss-diff} provides a model independent bound on the difference of errors for a classifier under distribution shift. By utilizing this bound, we can bound the difference of errors for the classifies generated by \gft~ algorithm under the distribution shift. 
\begin{equation*}
    \begin{array}{lll}
    \epsilon_2(\hat{h}_2)-\epsilon_1(\hat{h}_1) &\leq& \epsilon_2(\hat{h}_2)-\epsilon_2(\hat{h}_1)+ L\sqrt{R^2+1}W_p(D_1,D_2)  \\
    &\leq & \hat{\epsilon}_2(\hat{h}_2)-\hat{\epsilon}_2(\hat{h}_1) +4\sqrt{2}L\mathcal{R}_{n_2}(\mathcal{F})+4B\sqrt{\frac{\log{1/\delta}}{2n_2}}+L\sqrt{R^2+1}W_p(D_1,D_2)\\
    &\leq& 4\sqrt{2}L\mathcal{R}_{n_2}(\mathcal{F})+4B\sqrt{\frac{\log{1/\delta}}{2n_2}}+L\sqrt{R^2+1}W_p(D_1,D_2)
    \end{array}
\end{equation*}
where the third inequality hold as $\hat{h}_2$ is the minimizer of the empirical loss $\epsilon_2$. By iteratively applying this result, we have for any $t\in \{1,...,\kappa\}$
\begin{equation*}
    \begin{array}{lll}
    \epsilon_t(\hat{h}_\kappa)-\epsilon_1(\hat{h}_1) &\leq&
    \epsilon_\kappa(\hat{h}_\kappa)-\epsilon_1(\hat{h}_1) + L\sqrt{R^2+1}W_p(D_t,D_\kappa)\\
    &\leq &
    \sum_{i=1}^{\kappa-1}\left(\epsilon_{i+1}(\hat{h}_{i+1})-\epsilon_{i}(\hat{h}_{i})\right)+L\sqrt{R^2+1}\sum_{i=t}^{\kappa-1} W_p(D_i,D_{i+1})\\
     &\leq & \sum_{i=1}^{\kappa-1}\left[4\sqrt{2}L\mathcal{R}_{n_{i+1}}(\mathcal{F})+4B\sqrt{\frac{\log{(1/\delta)}}{2n_{i+1}}}+L\sqrt{R^2+1}W_p(D_i,D_{i+1})\right]+L\sqrt{R^2+1}W_p(D_t,D_\kappa).
    \end{array}
\end{equation*}
Summarize over $t=2,...,\kappa$, we have
\begin{equation*}
    \begin{array}{lll}
    &&\sum_{t=2}^\kappa \epsilon_t(\hat{h}_\kappa) - (\kappa - 1)\epsilon_1(\hat{h}_1)\\
    &\leq&
    4\sqrt{2}L(\kappa-1)\sum_{i=1}^{\kappa-1}\mathcal{R}_{n_{i+1}}(\mathcal{F})
    +4B(\kappa-1)\sum_{i=1}^{\kappa-1}\sqrt{\frac{\log{(1/\delta)}}{2n_{i+1}}}\\
    &&+L\sqrt{R^2+1}(\kappa - 1)\sum_{i=1}^{\kappa-1}W_p(D_i,D_{i+1})
    +L\sqrt{R^2+1}\sum_{t=2}^{\kappa}W_p(D_t,D_\kappa)
    \\
         & 
    \end{array}
\end{equation*}

Now, we present proofs for results in the theoretical analysis section. 
\begin{proof}
In this proof, we follow the same line as \cite{Wang2022UnderstandingGD}. By applying Corollary 2 of \cite{DBLP:journals/amai/KuznetsovM20}, we can bound the population loss of the classifier $\hat{h}_K$ from the final stage of GFT in the target domain $D_T$ as
\begin{equation*}
\begin{array}{lll}
    \epsilon_T(\hat{h}_\kappa)&\leq&\sum_{t=1}^\kappa\sum_{i=1}^{n_t}q_t^i\epsilon_t(\hat{h}_\kappa)+\disc(\mathbf{q_{\kappa}})+\|\mathbf{q}_{\kappa}\|_2\\
    &&+6B\sqrt{4\pi\log{\sum_{t=1}^\kappa n_t}}\mathcal{R}_{\kappa}^{seq}(\mathcal{H})+B\|\mathbf{q}_{\kappa}\|_{2}\sqrt{8\log{1/\delta}},
\end{array}
\end{equation*}
where $\mathcal{R}_{\kappa}^{\mathrm{seq}}$ is the sequential Rademacher complexity of the hypothesis space $\mathcal{H}$ with loss function $\mathcal{L}$.

By setting the optimal weights for discrepancy measurement as $\mathbf{q_{\kappa}}=\left(\frac{1}{n_1\kappa},...\frac{1}{n_1\kappa},...,\frac{1}{n_\kappa\kappa},...,\frac{1}{n_\kappa\kappa}\right)$, we can bound the $L^2$ norm of $\mathbf{q}_\kappa$ as
\begin{equation*}
    \|\mathbf{q}_{\kappa}\|_2 \leq \frac{1}{\kappa}\sqrt{\frac{1}{n_1}+\frac{1}{n_2}+...+\frac{1}{n_\kappa}}.
\end{equation*}

Then by applying the same weights, we can bound the discrepancy measurement as
\begin{equation*}
    \begin{array}{lll}
    \disc(\mathbf{q}_\kappa) &=& \sup_{h\in\mathcal{H}}\left(\epsilon_\kappa(h)-\sum_{t=1}^\kappa\sum_{i=1}^{n_t}q_i^t\epsilon_t(h)\right)  \\
        &\leq & \sup_{h\in\mathcal{H}}\left(\sum_{t=1}^\kappa\sum_{i=1}^{n_t}q_i^t|\epsilon_{\kappa}(h)-\epsilon_t(h)|\right) \\
    & \leq & \frac{L}{\kappa}\sqrt{R^2+1}\sum_{t=1}^{\kappa-1}W_p(D_t,D_\kappa)
    \end{array}
\end{equation*}
Finally, we provide the bound for the first term as
\begin{equation*}
\begin{array}{lll}
    \frac{1}{\kappa}\sum_{t=1}^\kappa\epsilon_t(\hat{h}_t) &\leq & \epsilon_1(\hat{h}_1) + \frac{4\sqrt{2L}(\kappa-1)}{\kappa}\sum_{t=1}^{\kappa-1}\mathcal{R}_{n_{t+1}}
    +  \frac{4B(\kappa-1)}{\kappa}\sum_{t=1}^{\kappa-1}\sqrt{\frac{\log{(1/\delta)}}{2n_{t+1}}}\\
    & & +
    \frac{L\sqrt{R^2+1}}{\kappa}\sum_{t=1}^{\kappa-1}W_p(D_t,D_{t+1})+
    \frac{L\sqrt{R^2+1}}{\kappa}\sum_{t=1}^{\kappa-1}W_p(D_t,D_\kappa)
\end{array}
\end{equation*}

Then, we can rewrite the above result as
\begin{equation*}
\begin{array}{l l l}
    \epsilon_\kappa(\hat{h}_\kappa) &\leq&\epsilon_1(\hat{h}_1) 
    +\frac{\kappa-1}{\kappa}L\sqrt{R^2+1}\Delta + (\kappa-1)L\sqrt{R^2+1}\Delta\\
    &&+\frac{4\sqrt{2}L(\kappa-1)}{\kappa}\sum_{t=1}^{\kappa-1}\mathcal{R}_{n_{t+1}}+\frac{4B(\kappa-1)}{\kappa}\sum_{t=1}^{\kappa-1}\sqrt{\frac{\log(1/\delta)}{2n_{t+1}}}\\
    &&+\frac{1}{\kappa}\sqrt{\sum_{t=1}^{\kappa}\frac{1}{n_t}}+6B\sqrt{4\pi\log{\sum_{t=1}^\kappa}n_t}\mathcal{R}_\kappa^{seq}(\mathcal{H})+\frac{B}{\kappa}\sqrt{8\log{(1/\delta)}\sum_{t=1}^\kappa\frac{1}{n_t}}
\end{array}
\end{equation*}
Furthermore, we expand $\epsilon_1{\hat{h}_1}$ and obtain the following result
\begin{equation*}
\begin{array}{lll}
    \epsilon_\kappa(\hat{h}_\kappa)&\leq&\hat{\epsilon}_1(\hat{h}_1)
    +\frac{2\sqrt{2}BL}{\sqrt{n_1}}+2B\sqrt{\frac{\log{(1/\delta)}}{2n_1}}+ (\kappa-\frac{1}{\kappa})L\sqrt{R^2+1}\Delta\\
    &&+\frac{4\sqrt{2}LB(\kappa-1)}{\kappa}\sum_{t=1}^{\kappa-1}\frac{1}{\sqrt{n_{t+1}}}+\frac{4B(\kappa-1)}{\kappa}\sum_{t=1}^{\kappa-1}\sqrt{\frac{\log(1/\delta)}{2n_{t+1}}}\\
    &&+\frac{1}{\kappa}\sqrt{\sum_{t=1}^{\kappa}\frac{1}{n_t}}+6B\sqrt{4\pi\log{\sum_{t=1}^\kappa}n_t}\mathcal{R}_\kappa^{seq}(\mathcal{H})+\frac{B}{\kappa}\sqrt{8\log{(1/\delta)}\sum_{t=1}^\kappa\frac{1}{n_t}}
\end{array}
\end{equation*}
which concludes the proof.
\end{proof}

For comparison, we also provide the expected error bounds for two baselines. The first one is training with the joint of data from all source domains. The second baseline is training a model only on the closest domain.

\section{Simulation Results}\label{app:sim-res}

\begin{figure}[t]
\centering
\subfigure[Classifier trained on dataset 1]{
    \includegraphics[width=0.95\textwidth]{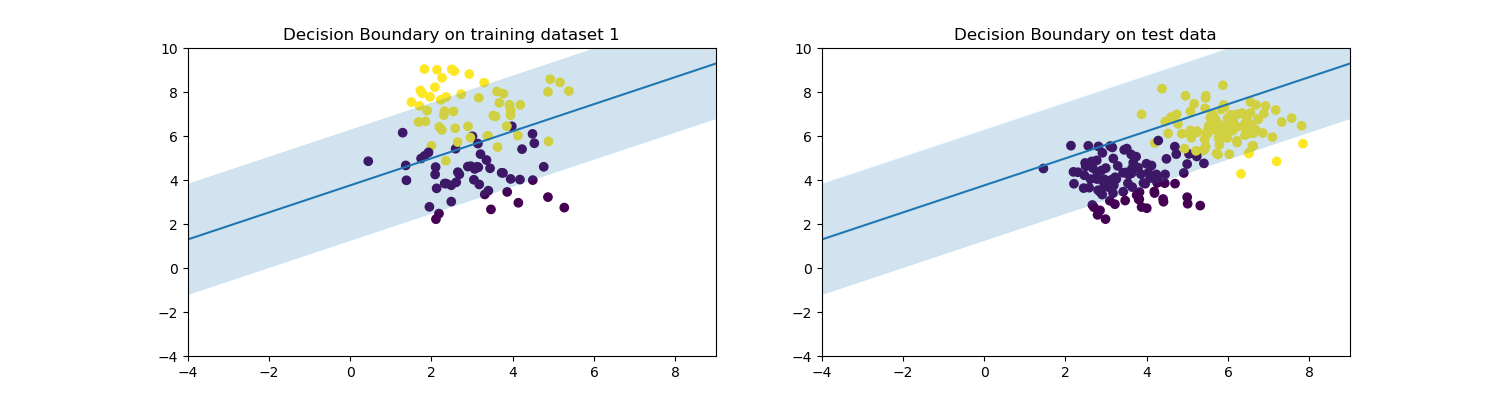}}
\hfill
\subfigure[Classifier trained on dataset 2]{
    \includegraphics[width=0.95\textwidth]{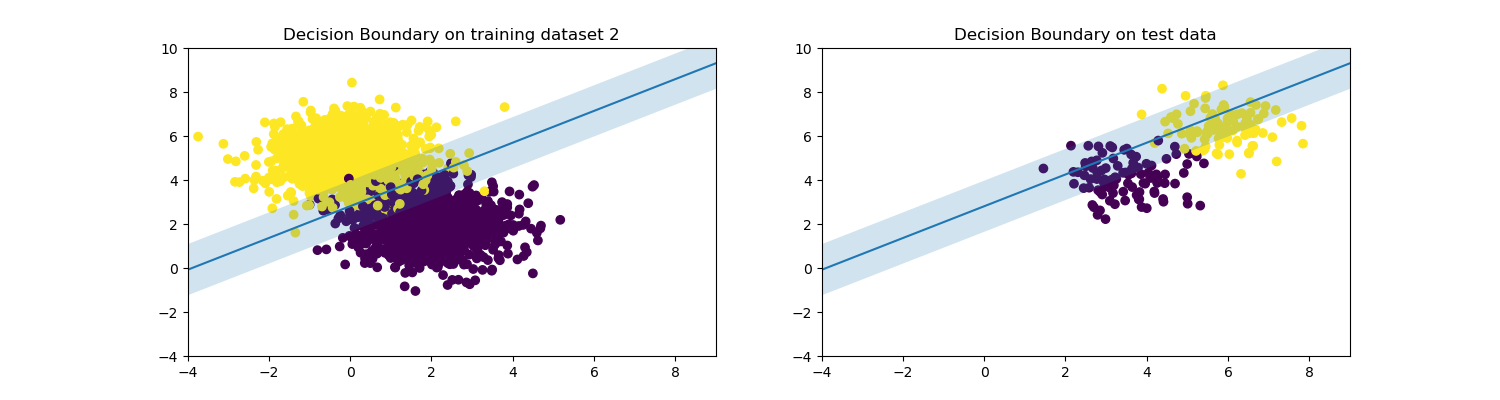}}
\hfill
\subfigure[Classifier trained on joint of datasets]{
    \includegraphics[width=0.95\textwidth]{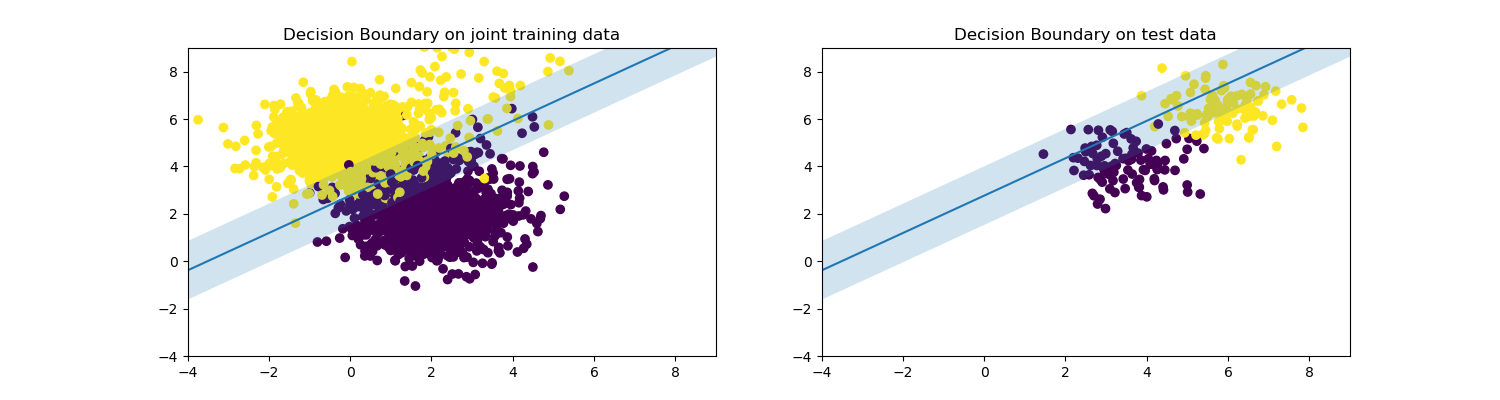}}
\hfill
\subfigure[Classfier trained with GFT]{
    \includegraphics[width=0.95\textwidth]{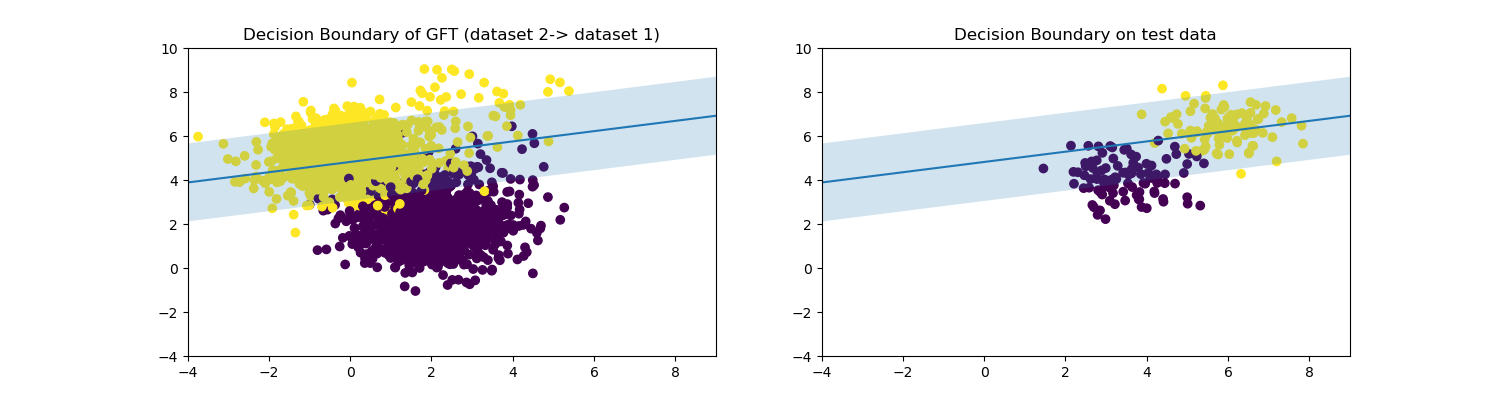}}
\caption{Two datasets drawn from different distributions are available as training data. The test data from target domain has higher discrepancy to dataset 1 than dataset 2. An linear model trained on dataset 1 achieves $0.555$ accuracy on test data as shown in subfigure (a). The same model trained only on dataset 2 achieves $0.54$ accuracy on test data. Although dataset 1 is more similar to the test dataset, it still achieves performance since the number of samples is very limited. By jointing the two sources, the model's accuracy is $0.53$ as dataset 2 has much more samples than 1. In the last subfigure, we applied GFT algorithm by following the order Source $2 \to$ Source 1. The modeled achieves 0.805 accuracy.}
\label{fig:sim-result}
\end{figure}

We present the results of the 2-source initial simulation experiments for our \gft~framework. We compared three training strategies: 1) We train two linear classifiers on Source $1$ and Source $2$, separately. 2) We train a single classifier on joint of Source $1$ and Source $2$. 3) We apply \gft~following path ``Source $2\to$ Source $1$''. The result is shown in Figure \ref{fig:sim-result}. \gft~achieves the highest test accuracy among the three strategies. The intuition is the classifier in training strategy 3 achieve a better result by following the guidance of our graph routing \gft.

\section{Baseline: All Sources}\label{app:baseline-as}
One important baseline is the risk of the classifier trained with data from all sources. Let's denote the classifier that minimize the empirical loss over all samples as 
\begin{equation*}
    \hat{h}=\argmin_{h\in\mathcal{H}}\frac{1}{n_{1:K}}\sum_{t=1}^K\sum_{(x,y)\in S_t}\mathcal{L}(h(x),y).
\end{equation*}
By applying the same theory from \cite{DBLP:journals/amai/KuznetsovM20}, we are able to write the upper bound of the risk of $\hat{h}$ on the target domain.

\begin{lemma}
Suppose at each time step, a sample is i.i.d.~drawn from the entire training data-set. With probability $1-\delta$, the expected error of classifier $\hat{h}$ is bounded as
\begin{equation*}
\begin{array}{l l l}
     \epsilon_T(\hat{h}) &\leq&
     L\sqrt{R^2+1}\sum_{t=1}^K\left(\frac{n_t}{n_{1:K}}\right)W_P(D_t,D_T)\\
     &&+\sum_{t=1}^K\left(\frac{n_t}{n_{1:K}}\right)\hat{\epsilon}_t(\hat{h})+
     \sum_{t=1}^K\frac{\sqrt{n_t}B}{n_{1:K}}\\
     &&+\sum_{t=1}^K\frac{\log{(1/\delta)}\sqrt{n_t}}{n_{1:K}}
\end{array}
\end{equation*}
\end{lemma}

The expected error of $\hat{h}$ scales as the weighted Wasserstein-1 distance between each source and the target. When a domain has dominate number of sample $n_i$ and large enough $\Delta_{i,T}$, the error on this domain will dominate the final trained classifier.

\section{Baseline: Closest Source}\label{app:baseline-cs}
As in most domain adaption algorithms, training on the closest source has been shown to achieve good performance in many real applications. But the drawback of limited number of samples always exists when the closest source does not contain enough number of labeled training samples. Here, we denote the classifier trained on the closest domain $c$ as $\hat{h}_c=\arg\min_{h\in\mathcal{H}}\frac{1}{n_c}\sum_{i=1}^{n_c}\mathcal{L}(h(x),y)$.

\begin{lemma}
With probability $1-\delta$, the expected error of the learned classifier $\hat{h}_c$ satisfies
\begin{equation*}
\begin{array}{lll}
     &&\epsilon_T(\hat{h}_c) \leq L\sqrt{R^2+1}W_p(D_c,D_T)\\
     &&+\hat{\epsilon}_c(\hat{h}_c)+\frac{B}{\sqrt{n_{c}}}+\frac{\log{(1/\delta)}}{\sqrt{n_c}}
\end{array}
\end{equation*}
\end{lemma}
As in standard learning theorem, the risk decreases monotonically as the number of sample $n_c$ grows. In the case of $n_c$ does not have contain enough samples, a trade-off between $\Delta_{c,T}$ and $n_c$ need to be carefully considered and selected.

\section{Pairwise Distance Matrix}
\begin{figure}[t]
    \centering
    \begin{minipage}{0.45\textwidth}
    \centering
    \includegraphics[width=0.9\textwidth]{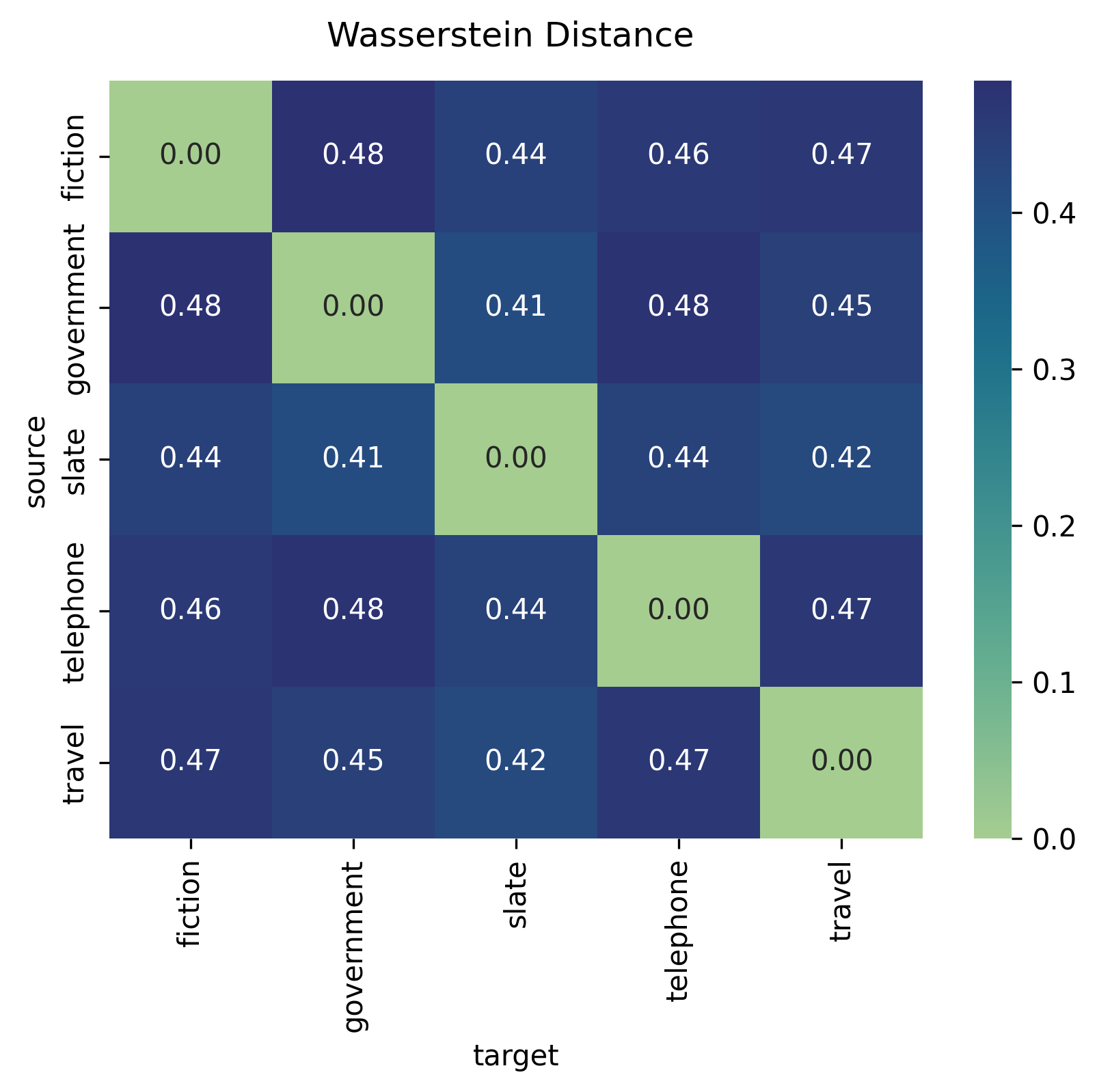}
    \caption{The matrix of pairwise Wasserstein-1 distance between domains in the \texttt{MultiNLI} dataset.}
    \label{fig:domain_distance_nli}
    \end{minipage}\hfill
    \begin{minipage}{0.45\textwidth}
    \centering
    \includegraphics[width=\textwidth]{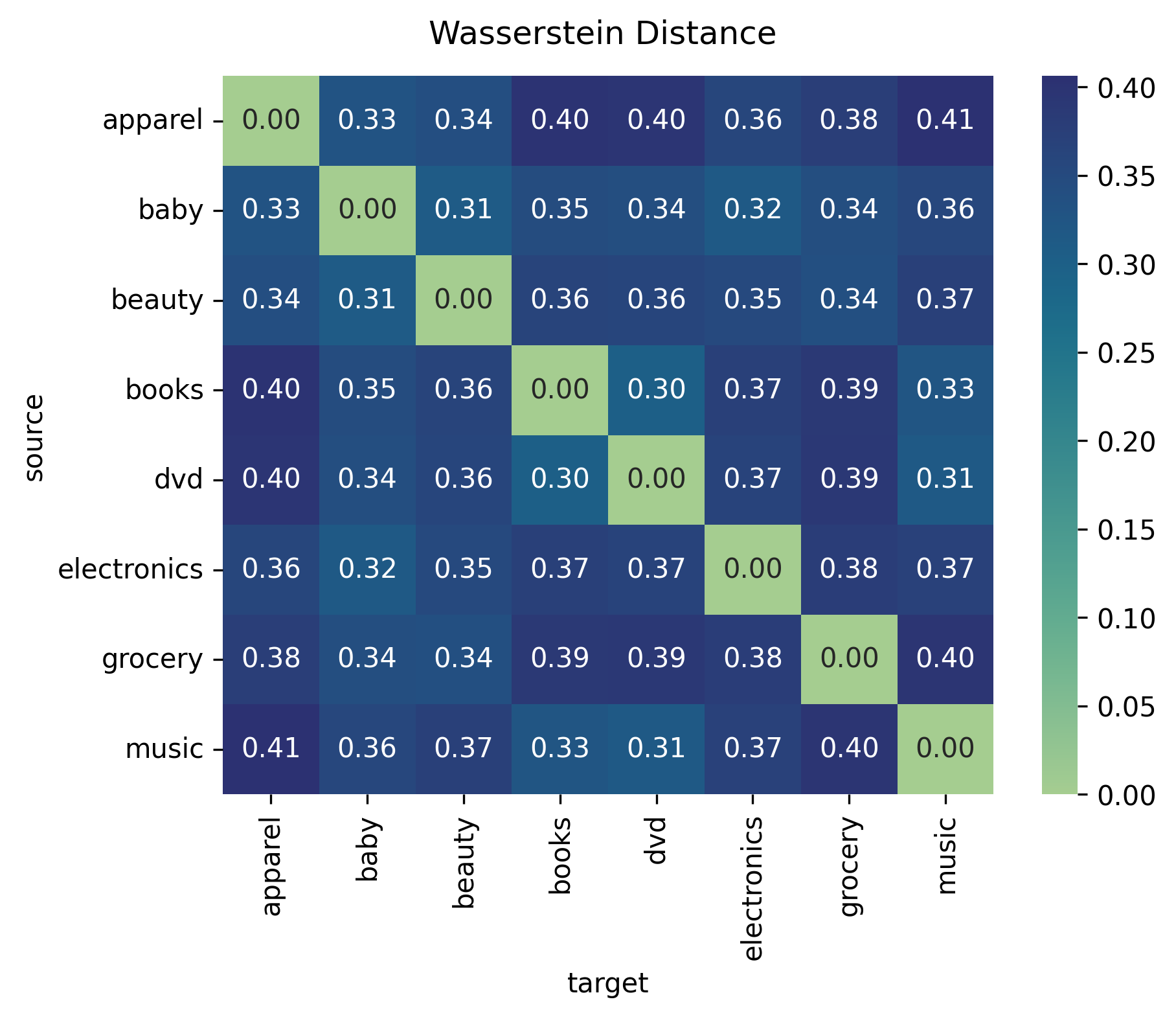} 
    \caption{The matrix of pairwise Wasserstein-1 distance between domains in  the Amazon dataset.}
    \label{fig:domain_distance_sa}
    \end{minipage}
\end{figure}

We include the $W_1$ adjacency matrix of \texttt{MultiNLI} and Amazon (for sentiment analysis) domains in Figure \ref{fig:domain_distance_nli} and \ref{fig:domain_distance_sa} respectively.
\label{appendix:pairwise_distance}

\end{document}